\documentclass[10pt,twocolumn,letterpaper]{article}

\usepackage[pagenumbers]{cvpr} 

\usepackage{booktabs}
\usepackage{multirow}
\usepackage{graphicx}  
\usepackage{array}
\usepackage{colortbl}
\usepackage{xcolor}
\usepackage{tcolorbox}
\usepackage{caption}
\usepackage{float}
\usepackage{placeins}
\tcbuselibrary{breakable, skins}

\definecolor{cvprblue}{rgb}{0.21,0.49,0.74}
\usepackage[pagebackref,breaklinks,colorlinks,allcolors=cvprblue]{hyperref}

\usepackage[dvipsnames]{xcolor}

\def\paperID{*****} 
\def\confName{CVPR}
\def\confYear{2026}

\title{Can Vision Language Models Judge Action Quality? An Empirical Evaluation}

\author{
Miguel Monte e Freitas$^{1,2}$ \quad Rui Henriques$^{2,3}$ \quad Ricardo Rei$^{1}$ \quad Pedro Henrique Martins$^{1}$ \\[6pt]
{\small $^{1}$Sword Health\quad $^{2}$Instituto Superior Técnico, Universidade de Lisboa\quad $^{3}$INESC-ID}\\[3.5pt] {\tt\small mm.freitas@swordhealth.com\quad ph.martins@swordhealth.com}}

\begin{document}

\maketitle
\begin{abstract}
Action Quality Assessment (AQA) has broad applications in physical therapy, sports coaching, and competitive judging. Although Vision Language Models (VLMs) hold considerable promise for AQA, their actual performance in this domain remains largely uncharacterised. We present a comprehensive evaluation of state-of-the-art VLMs across activity domains (e.g. fitness, figure skating, diving), tasks, representations, and prompting strategies. Baseline results reveal that Gemini 3.1 Pro, Qwen3-VL and InternVL3.5 models perform only marginally above random chance, and although strategies such as incorporation of skeleton information, grounding instructions, reasoning structures and in-context learning lead to isolated gains, none is consistently effective. Analysis of prediction distributions uncovers two systematic biases: a tendency to predict correct execution regardless of visual evidence, and a sensitivity to superficial linguistic framing. Reformulating tasks contrastively to mitigate these biases yields minimal improvement, suggesting that the models' limitations go beyond these biases, pointing to a fundamental difficulty with fine-grained movement quality assessment. Our findings establish a rigorous baseline for future VLM-based AQA research and provide an actionable outline for failure modes requiring mitigation prior to reliable real-world deployment.
\end{abstract}    
\section{Introduction}
\label{sec:intro}

Action Quality Assessment (AQA) is the field of computer vision focused on the automated evaluation of action execution quality. It has numerous high-impact applications: in physical therapy, it can deliver real-time feedback to patients following rehabilitation protocols without requiring human supervision \cite{capecci2019kimore, vakanski2018dataset}; in recreational sports, it can assess an athlete's form to drive improvement and prevent injury-prone technical flaws \cite{parmar2022domain}; in competitive judging, it can serve as a more objective and consistent judge \cite{parmar2017learning}.

AQA has traditionally been framed as a regression task, where the goal is to predict a scalar quality score \cite{parmar2017learning, carreira2017quo, xiang2018s3d, dong2021learning, zeng2020hybrid, wang2021tsa, nagai2021action, vaswani2017attention, xu2022likert, iyer2022action, bai2022action, gedamu2023finegrained, fang2023end}. While significant progress has been made within this framing, predicted scores offer little actionable insight or explainability, limiting their practical value. To address these shortcomings, recent work has begun exploring textual feedback generation as a richer alternative \cite{parmar2019what, zhang2024narrative, li2025techcoach}.

In parallel, Vision Language Models (VLMs) have made significant progress in generating text conditioned on both image and video inputs. Unlike more specialised models, VLMs possess broad prior knowledge that often enables them to generalise to new tasks without task-specific training. For AQA, they present three theoretically promising advantages: (1) their prompt-based interface makes it straightforward to incorporate additional information or adjust output formats without retraining; (2) their conversational nature enables interactive applications, where users could follow-up on generated feedback with specific questions; and (3) reasoning VLMs (e.g., Chain-of-Thought \cite{wei2022chain}) provide some transparency into the factors driving their assessments through reasoning traces, which would increase the explainability of assessments.

Recent studies have started to explore the role of VLMs in AQA \cite{zhang2024narrative, wu2025hieroaction}, hinting at their potential. Yet, these efforts have been limited to specific tasks and datasets, leaving a key question unanswered: can off-the-shelf VLMs reliably perform AQA across diverse human activities and tasks?

To answer this question, we evaluate state-of-the-art models from the Gemini 3 \cite{googleDeepMind2026gemini31procard}, Qwen3-VL \cite{qwen2025qwen3vl}, and InternVL3.5 \cite{wang2025internvl35} families on established AQA datasets, covering multiple domains (\textit{bodyweight} and \textit{weighted exercises}, \textit{competitive diving}, and \textit{competitive figure skating}) and tasks (\textit{visual question answering}, \textit{error detection}, \textit{technical guideline verification}, and \textit{score regression}).

Since skeleton representations have been commonly used in movement analysis \cite{qiu2022poseguided, chen2021sportscap, pan2019action, nekoui2021eagleeye, baptista2021human}, we investigate their integration in VLM-based AQA through different visual preprocessing methods. We further explore the effect of different prompting strategies, including grounding instructions, templated reasoning structures, technical guidelines, and in-context learning \cite{brown2020language}, with the aim of improving test-time performance without additional training. We also examine how susceptible VLMs are to linguistic biases by measuring the effect of prompt phrasing variations on model predictions, and explore contrastive task reformulation to mitigate these biases.

Results across five tasks, three preprocessing methods, and seven prompting strategies consistently show that VLMs struggle with AQA, often performing at near-random levels. Although some preprocessing and prompting strategies yield modest gains in isolated cases, such as cropping and in-context learning, none proves consistently effective.

Beyond raw performance, our analysis reveals two systematic biases in VLMs: a tendency to predict correct execution regardless of visual content, and a sensitivity to superficial linguistic cues. Attempts to mitigate these biases through contrastive task reformulation still yield poor performance, suggesting that VLMs face fundamental limitations in movement quality assessment that extend beyond these biases.

This comprehensive evaluation offers a reference benchmark and establishes a rigorous baseline for future VLM-based AQA research, while providing an actionable outline of the current failure modes that the field must address.
\section{Methodology}
\label{sec:datasets_and_tasks}

We evaluate state-of-the-art VLMs on AQA across a diverse set of datasets and tasks. This section describes the datasets and tasks used for evaluation (Section~\ref{sec:datasets_and_tasks}), the models selected for comparison (Section~\ref{sec:models}), the evaluation metrics adopted (Section~\ref{sec:metrics}), and the implementation details of our experimental setup (Section~\ref{sec:implementation}).

\subsection{Datasets and Tasks}
\label{sec:datasets_and_tasks}

A selected range of activites, data modalities and tasks is proposed for a thorough assessment of VLMs' performance in AQA. We include three gym datasets covering both bodyweight and weight-loaded exercises, alongside two Olympic sport datasets, namely diving and figure skating. Regarding modalities, we consider RGB videos and individual keyframes. The tasks covered include VQA, technical keypoint verification, error detection, and action quality score regression. The full prompts used for all tasks are provided in Appendix \ref{sec:appendix_prompts}.

\subsubsection{LLM-FMS}
LLM-FMS \cite{xing2025llmfms} is a keyframe-based dataset covering 7 exercises from the Functional Movement Screen protocol, containing 1,812 keyframes from 45 subjects. Each keyframe is paired with exercise-specific VQA questions and answer options. We prompt the model with a keyframe and its corresponding questions, asking it to return answers as a JSON object.

\subsubsection{EgoExo-Fitness}
EgoExo-Fitness \cite{li2024egoexo} contains 913 annotated fitness video instances across 12 exercise types, recorded from synchronised egocentric and exocentric cameras. Each instance is labelled for adherence to specific technical guidelines (\textit{e.g.}, ``\textit{Keep your back straight}"). We prompt the model with a frontal-view video and a single guideline, asking it to predict whether the subject complies.

\subsubsection{Fitness-AQA}
Fitness-AQA \cite{parmar2022domain} covers three gym exercises — barbell rows, overhead press, and squat — each with two annotated form errors. We evaluate on the overhead press (339 videos) and squat (224 videos) test sets, prompting the model with a video and error descriptions and asking it to output a JSON object indicating which errors are present.

\subsubsection{FineFS}
FineFS \cite{ji2023localization} consists of figure skating competition videos, each showing an athlete performing a series of elements (jumps, spins, or sequences) annotated with fine-grained scores including the Grade of Execution (GOE), ranging from -5 to 5. We randomly sample 500 elements stratified by type and prompt the model to predict the GOE from each element's video clip. As scoreboards displaying the ground truth GOE are visible in the footage, we mask them prior to evaluation to prevent data leakage.

\subsubsection{MTL-AQA}
MTL-AQA \cite{parmar2019what} contains 353 diving competition videos, each annotated with a difficulty score and a final score from a panel of 5 or 7 judges. We use these annotations to derive the average judge execution score for each dive, ranging from 0 to 10, and prompt the model to predict this value from the dive video.

\subsection{Models}
\label{sec:models}

We evaluate models from three state-of-the-art VLM families: Gemini 3 \cite{googleDeepMind2026gemini31procard}, Qwen3-VL \cite{qwen2025qwen3vl}, and InternVL3.5 \cite{wang2025internvl35}. For Gemini, we use Gemini 3.1 Pro Preview, their most capable model to date, configured with high thinking level. For Qwen3-VL, we select the largest models from both the instruct and reasoning paradigms: Qwen3-VL-235B-A22B-Instruct and Qwen3-VL-235B-A22B-Thinking. For InternVL3.5, we use their flagship model, InternVL3.5-241B-A28B, evaluated in both standard and thinking modes.
Collectively, these models span closed- and open-source options across instruct and thinking paradigms, providing broad coverage of the current state of the art.

\subsection{Evaluation Metrics}
\label{sec:metrics}

To evaluate the VLMs in classification tasks, we use balanced accuracy, defined as the average recall across classes,
\begin{equation}
\small
    \text{Balanced Accuracy} = \frac{1}{C} \sum_{c=1}^{C} \frac{\text{TP}_c}{\text{TP}_c + \text{FN}_c},
\end{equation}
where $C$ is the number of classes, $\text{TP}_c$ is the number of true positives for class $c$, and $\text{FN}_c$ is the number of false negatives for class $c$. Balanced accuracy is computed independently for each exercise and then averaged across exercises, ensuring that exercises with more samples do not disproportionately influence the overall results.

For regression tasks, we adopt Spearman rank correlation and relative $\ell_2$ distance, both standard metrics in AQA \cite{yin2025decade}. The Spearman correlation measures the monotonic relationship between predicted and ground-truth scores based on the differences between their respective ranks. Relative $\ell_2$ distance measures the magnitude of prediction error relative to the range of ground-truth scores,
\begin{equation}
\small
    \text{R-}\ell_2 = \frac{\sqrt{\frac{1}{N}\sum_{i=1}^{N}(\hat{y}_i - y_i)^2}}{\max(y) - \min(y)},
\end{equation}
where $\hat{y}_i$ and $y_i$ are the predicted and ground-truth scores for the $i$-th sample, respectively.

\subsection{Implementation Details}
\label{sec:implementation}

Open-source models are served using vLLM \cite{kwon2023efficient}, while Gemini is accessed through Vertex AI \cite{vertexai2023}. All models are used with their recommended sampling configurations: Gemini 3.1 Pro Preview is run with a temperature of 1.0, top-$p$ of 0.95, top-$k$ of 64, and a candidate count of 1; Qwen3-VL-235B-A22B-Instruct with a temperature of 0.7, top-$p$ of 0.8, top-$k$ of 20, and a repetition penalty of 1.0, while its thinking variant uses the same settings except with top-$p$ raised to 0.95. As for InternVL3.5-241B-A28B no optimal configuration has been publicly disclosed, we adopt the same parameters as Qwen3-VL-235B-A22B-Instruct, and reduce the temperature to 0.6 for the thinking mode, as advised by OpenGVLab.

Due to InternVL's shorter context window, samples from EgoExo-Fitness, Fitness-AQA, and the sequence class of FineFS cannot be processed at their original resolution. For these samples, spatial resolution is downsampled to 50\% of its original value, and videos exceeding 120 frames are uniformly subsampled to this limit.
\section{Experiments}
\label{sec:results}

\subsection{Baseline results}
\begin{table*}[t]
\centering
\renewcommand{\arraystretch}{0.85}
\setlength{\tabcolsep}{4pt}
\scriptsize
\begin{tabular*}{\textwidth}{@{} l @{\extracolsep{\fill}} c c c c c c c @{}}
\toprule
& \multicolumn{3}{c}{\textbf{Classification}} & \multicolumn{4}{c}{\textbf{Regression}} \\
\cmidrule(lr){2-4} \cmidrule(l){5-8}
& \textbf{LLM-FMS} & \textbf{EgoExo-Fitness} & \textbf{Fitness-AQA} & \multicolumn{2}{c}{\textbf{FineFS}} & \multicolumn{2}{c}{\textbf{MTL-AQA}} \\
\noalign{\vspace{3pt}}
\textbf{Model} & Bal.\ Acc.$\uparrow$ & Bal.\ Acc.$\uparrow$ & Bal.\ Acc.$\uparrow$ & $\rho\uparrow$ & $\text{R-}\ell_2\downarrow$ & $\rho\uparrow$ & $\text{R-}\ell_2\downarrow$ \\
\midrule
Random guess          & 0.4310          & 0.5000          & 0.5000          & 0.0000          & 0.9312          & 0.0000          & 0.2181 \\
\midrule
Qwen3-VL-Instruct     & 0.4725          & \textbf{0.5733} & 0.5557 & 0.1998          & 0.2590          & 0.2319          & 0.2026 \\
Qwen3-VL-Thinking     & 0.4605          & 0.5650          & 0.5224          & \textbf{0.2806}          & 0.2673          & 0.1367          & 0.2665 \\
InternVL3.5           & 0.4488          & 0.5429          & 0.5296          & 0.2042          & 0.2540          & \textbf{0.2636} & \textbf{0.1847} \\
InternVL3.5$^\dagger$ & 0.4697          & 0.5461          & 0.5131          & 0.2797          & \textbf{0.2372} & 0.1974          & 0.2487 \\
Gemini 3.1 Pro        & \textbf{0.6029} & 0.5167          & \textbf{0.5596}          & 0.2651 & 0.3079          & 0.0576          & 0.3060 \\
\bottomrule
\multicolumn{8}{@{}l}{$^\dagger$ thinking mode.}
\end{tabular*}
\caption{\textbf{Baseline results.} \textbf{Classification} tasks (LLM-FMS, EgoExo-Fitness, Fitness-AQA) are evaluated with balanced accuracy. \textbf{Regression} tasks (FineFS, MTL-AQA) are evaluated with Spearman correlation ($\rho$) and R-$\ell_2$. Best result per column shown in \textbf{bold}. Random guess represents uniform sampling across labels for classification, and predicting the ground truth mean for regression.}
\label{tab:baseline_results}
\end{table*}

Baseline experiments on classification tasks (Table \ref{tab:baseline_results}) show consistently poor performance across the different models, with little improvement over random guess. Qwen3-VL-Instruct achieves slightly higher balanced accuracy on the dataset with the longest videos (EgoExo-Fitness), while Gemini 3.1 Pro performs considerably better on the image dataset (LLM-FMS).

For regression tasks, the results show positive rank correlations with ground truth scores, although still weak. Performance is inconsistent across datasets: thinking models perform considerably better on FineFS than on MTL-AQA, whereas instruct models show the opposite pattern.

\subsection{Incorporation of Skeleton Data}

Although VLMs can process raw RGB videos without task-specific preprocessing, it remains an open question whether additional visual preprocessing could benefit these models for AQA. Since AQA requires attention to the subject's body position and movement, preprocessing that makes the subject and their pose more visually salient could improve performance. Under this hypothesis, three preprocessing methods are explored (see Appendix \ref{sec:appendix_datasets} for examples):
\begin{itemize} 
    \item \textbf{Cropped:} Frames are cropped to include only the subject, eliminating uninformative content. The bounding box (used to crop the frame) is defined from the skeleton joints with a small margin.
    \item \textbf{Skeleton overlays:} Skeleton joints and their connections are drawn on top of the original frames, making the key body landmarks more prominent.
    \item \textbf{Skeleton-only:} Frames are replaced by skeleton joints and connections rendered on a plain white background, isolating body structure and movement\footnote{All skeletons are estimated using the SAM 3D Body model \cite{yang2025sam3dbody} and follow the standard COCO 17 keypoint format.}.
\end{itemize}

\begin{table}[t]
\centering
\renewcommand{\arraystretch}{0.85}
\setlength{\tabcolsep}{0pt}
\scriptsize
\begin{tabular*}{\columnwidth}{@{} l @{\extracolsep{\fill}} l c c c c @{\hspace{4pt}}}
\toprule
\textbf{Dataset} & \textbf{Model} & \textbf{Original} & \textbf{Cropped} & \textbf{Sk. overlays} & \textbf{Sk. only} \\
\midrule
\multirow{5}{*}{\rotatebox{90}{LLM-FMS}}
 & Qwen3-VL-Instruct        & 0.4725 & \textbf{0.4808} & 0.4600 & 0.4374 \\
 & Qwen3-VL-Thinking        & 0.4605 & 0.\textbf{4693} & 0.4678 & 0.4416 \\
 & InternVL3.5              & 0.4488 & \textbf{0.4762} & 0.4473 & 0.4443 \\
 & InternVL3.5$^\dagger$    & \textbf{0.4697} & 0.4591 & 0.4576 & 0.4628 \\
 & Gemini 3.1 Pro           & 0.6029 & 0.6257 & \textbf{0.6298} & 0.5925 \\
\midrule
\multirow{5}{*}{\rotatebox{90}{\shortstack{EgoExo-\\Fitness}}}
 & Qwen3-VL-Instruct        & \textbf{0.5733} & 0.5725 & 0.5472 & 0.5403 \\
 & Qwen3-VL-Thinking        & \textbf{0.5650} & 0.5613 & 0.5544 & 0.5370 \\
 & InternVL3.5              & 0.5429 & \textbf{0.5737} & 0.5708 & 0.5381 \\
 & InternVL3.5$^\dagger$    & \textbf{0.5461} & 0.5447 & 0.5386 & 0.5267 \\
 & Gemini 3.1 Pro           & 0.5167 & \textbf{0.6002} & 0.5825 & 0.5512     \\
\midrule
\multirow{5}{*}{\rotatebox{90}{\shortstack{Fitness-\\AQA}}}
 & Qwen3-VL-Instruct        & \textbf{0.5557} & 0.5505 & 0.5358 & 0.5333 \\
 & Qwen3-VL-Thinking        & 0.5224 & \textbf{0.5279} & 0.5075 & 0.4966 \\
 & InternVL3.5              & \textbf{0.5296} & 0.5243 & 0.5007 & 0.5059 \\
 & InternVL3.5$^\dagger$    & \textbf{0.5131} & 0.5068 & 0.5030 & 0.4874 \\
 & Gemini 3.1 Pro           & 0.5596 & 0.5631 & \textbf{0.5668} & 0.5523 \\
\bottomrule
\multicolumn{6}{@{}l}{$^\dagger$ thinking mode. Sk.: Skeleton.}
\end{tabular*}
\caption{Balanced accuracy (\%) across preprocessing methods for classification tasks, with best result per row shown in \textbf{bold}.}
\label{tab:skeleton_classification}
\end{table}
\begin{table}[t]
\centering
\renewcommand{\arraystretch}{0.85}
\setlength{\tabcolsep}{0pt}
\scriptsize
\begin{tabular*}{\columnwidth}{@{} l @{\extracolsep{\fill}} l @{\hspace{4pt}} c@{\hspace{2pt}}c @{\hspace{4pt}} c@{\hspace{2pt}}c @{\hspace{4pt}} c@{\hspace{2pt}}c @{\hspace{2pt}}}
\toprule
& & \multicolumn{2}{c@{\hspace{4pt}}}{\textbf{Original}} & \multicolumn{2}{c@{\hspace{4pt}}}{\textbf{Sk. overlays}} & \multicolumn{2}{c}{\textbf{Sk. only}} \\
\textbf{Dataset} & \textbf{Model} & $\rho\uparrow$ & $\text{R-}\ell_2\downarrow$ & $\rho\uparrow$ & $\text{R-}\ell_2\downarrow$  & $\rho\uparrow$ & $\text{R-}\ell_2\downarrow$ \\
\midrule
\multirow{5}{*}{\rotatebox{90}{FineFS}}
 & Qwen3-VL-Instruct        & 0.1998 & \textbf{0.2590} & \textbf{0.2041} & 0.2690 & 0.0507  & 0.3004 \\
 & Qwen3-VL-Thinking        & 0.2806 & 0.2673 & \textbf{0.3071} & 0.2909 & 0.0457  & \textbf{0.2412} \\
 & InternVL3.5              & 0.2042 & 0.2540 & \textbf{0.2687} & 0.2817 & -0.0020 & \textbf{0.2186} \\
 & InternVL3.5$^\dagger$    & 0.2797 & 0.2372 & \textbf{0.3110} & 0.2633 & 0.0751  & \textbf{0.1702} \\
 & Gemini 3.1 Pro           & \textbf{0.2651} & 0.3079 & 0.2435     & \textbf{0.3037}     & 0.0385  & 0.3626 \\
\midrule
\multirow{5}{*}{\rotatebox{90}{MTL-AQA}}
 & Qwen3-VL-Instruct        & \textbf{0.2319} & 0.2026 & 0.0716 & 0.2326 & -0.0361 & \textbf{0.1798} \\
 & Qwen3-VL-Thinking        & 0.1367 & 0.2665 & \textbf{0.1617} & 0.2836 & 0.0035  & \textbf{0.2278} \\
 & InternVL3.5              & \textbf{0.2636} & 0.1847 & 0.1265 & 0.2141 & 0.0337  & \textbf{0.1710} \\
 & InternVL3.5$^\dagger$    & \textbf{0.1974} & 0.2487 & 0.0987 & 0.2683 & 0.0226  & \textbf{0.1931} \\
 & Gemini 3.1 Pro           & 0.0576 & 0.3060 & \textbf{0.0769} & \textbf{0.2897} & 0.0479  & 0.4274 \\
\bottomrule
\multicolumn{8}{@{}l}{$^\dagger$ thinking mode. Sk.: Skeleton.}
\end{tabular*}
\caption{Spearman correlation ($\rho$) and $\text{R-}\ell_2$ across preprocessing methods for the regression tasks, with best result per row for each metric shown in \textbf{bold}. Cropping was not applied to these modalities, as it could omit important contextual factors such as jump height or splash.}
\label{tab:skeleton_regression}
\end{table}

\paragraph{Classification results.} As shown in Table \ref{tab:skeleton_classification}, cropping often yields small gains, though the effect is inconsistent across models. Direct incorporation of skeleton information, whether via overlays or skeleton-only renders, rarely improves performance.

\paragraph{Regression results.} On regression tasks (Table \ref{tab:skeleton_regression}), the skeleton overlays show some tendency to improve rank correlation scores, although not consistently. Skeleton-only renders, however, reduce correlation to near zero across all models and datasets. This likely reflects the complexity of movements in these datasets, which VLMs may struggle to interpret from skeleton representations alone. The notably lower $\text{R-}\ell_2$ values reflect the model defaulting to conservative, low-variance predictions that incidentally cluster closer to the ground truth mean.

\subsection{Prompt Engineering}

Prompt engineering is known to improve model performance by directing the model's attention and activating task-relevant knowledge \cite{schulhoff2024prompt}, making it an essential component of a comprehensive evaluation of VLMs on AQA. We experiment with the following prompting strategies (full prompts provided in Appendix \ref{sec:appendix_prompts}):
\begin{itemize}
    \item \textbf{Base prompt:} A minimal prompt specifying the task and the required output format, serving as a baseline.
    \item \textbf{Visual grounding:} Instructs the model to begin by thoroughly analysing the visual content before producing any output, anchoring its responses in direct observation.
    \item \textbf{Two-step observation:} Prompts the VLM to generate a detailed description of the visual content before producing an answer, encouraging more deliberate reasoning.
    \item \textbf{Guidelines:} Provides brief quality criteria giving the model domain-relevant reference points. We experiment with two variants: \textit{positive guidelines} describe what the exercise should look like when performed correctly, while \textit{negative guidelines} describe common errors made during the exercise (see Appendix \ref{sec:appendix_guidelines} for examples).
    \item \textbf{Structured reasoning:} Defines an explicit reasoning structure on the model's output. Inspired by the reasoning steps proposed by \citet{wu2025hieroaction}, the model is prompted to organise its reasoning into the following stages:
    \begin{itemize}
        \item \texttt{<look></look>}: A high-level description of the visual content.
        \item \texttt{<decompose></decompose>}: Identification of the specific components to be analysed (e.g., body parts, joint angles, movement speed).
        \item \texttt{<analyse></analyse>}: A focused analysis of each identified component.
        \item \texttt{<assess></assess>}: High-level reasoning synthesising the component analyses into an overall assessment.
        \item \texttt{<output></output>}: The final answer in the required format.
    \end{itemize}
    \item \textbf{In-Context Learning:} Provides the model with a set of input-output examples at inference time, often improving performance without any fine-tuning \cite{brown2020language}. For classification tasks, one example per output class is provided. For regression tasks, one high-score and one low-score example of the same action type is included. Examples are added as conversation history rather than in-prompt.
\end{itemize}

\begin{table*}[t]
\centering
\renewcommand{\arraystretch}{0.85}
\setlength{\tabcolsep}{0pt}
\scriptsize
\newcommand{\mcc}[1]{\multicolumn{2}{c@{\hspace{6pt}}}{#1}}  
\begin{tabular*}{\textwidth}{@{} l @{\extracolsep{\fill}} l
  @{\hspace{4pt}} c@{\hspace{2pt}}c
  @{\hspace{4pt}} c@{\hspace{2pt}}c
  @{\hspace{4pt}} c@{\hspace{2pt}}c
  @{\hspace{4pt}} c@{\hspace{2pt}}c
  @{\hspace{4pt}} c@{\hspace{2pt}}c
  @{\hspace{4pt}} c@{\hspace{2pt}}c
  @{\hspace{4pt}} c@{\hspace{2pt}}c @{\hspace{2pt}}}
\toprule
\textbf{Dataset} & \textbf{Model}
  & \multicolumn{2}{c@{\hspace{4pt}}}{\textbf{\shortstack{Base\\Prompt}}}
  & \multicolumn{2}{c@{\hspace{4pt}}}{\textbf{\shortstack{Visual\\Grounding}}}
  & \multicolumn{2}{c@{\hspace{4pt}}}{\textbf{\shortstack{Two-\\Step}}}
  & \multicolumn{2}{c@{\hspace{4pt}}}{\textbf{\shortstack{Structured\\Reasoning}}}
  & \multicolumn{2}{c@{\hspace{4pt}}}{\textbf{\shortstack{Positive\\Guidelines}}}
  & \multicolumn{2}{c@{\hspace{4pt}}}{\textbf{\shortstack{Negative\\Guidelines}}}
  & \multicolumn{2}{c}{\textbf{ICL}} \\
\midrule
\textit{Classification} &
  & \mcc{Bal.\ Acc.$\uparrow$}
  & \mcc{Bal.\ Acc.$\uparrow$}
  & \mcc{Bal.\ Acc.$\uparrow$}
  & \mcc{Bal.\ Acc.$\uparrow$}
  & \mcc{Bal.\ Acc.$\uparrow$}
  & \mcc{Bal.\ Acc.$\uparrow$}
  & \mcc{Bal.\ Acc.$\uparrow$} \\[2pt]
\multirow{5}{*}{\rotatebox{90}{LLM-FMS}}
 & Qwen3-VL-Instruct\vphantom{$^\dagger$}
   & \mcc{\textbf{0.4725}} & \mcc{0.4631} & \mcc{0.4634} & \mcc{0.4397} & \mcc{0.4659} & \mcc{0.4404} & \mcc{\textbf{0.5038}} \\
 & Qwen3-VL-Thinking\vphantom{$^\dagger$}
   & \mcc{0.4605} & \mcc{\textbf{0.4685}} & \mcc{0.4248} & \mcc{0.4397} & \mcc{0.4572} & \mcc{\textbf{0.4670}} & \mcc{0.4655} \\
 & InternVL3.5\vphantom{$^\dagger$}
   & \mcc{0.4488} & \mcc{0.4554} & \mcc{0.4446} & \mcc{\textbf{0.4640}} & \mcc{0.4518} & \mcc{0.4376} & \mcc{\textbf{0.4937}} \\
 & InternVL3.5$^\dagger$
   & \mcc{\textbf{0.4697}} & \mcc{0.4508} & \mcc{0.4572} & \mcc{0.4582} & \mcc{0.4549} & \mcc{0.4319} & \mcc{\textbf{0.4829}} \\
 & Gemini 3.1 Pro\vphantom{$^\dagger$}
   & \mcc{0.6029} & \mcc{0.6046} & \mcc{0.5913} & \mcc{0.5949} & \mcc{0.5788} & \mcc{\textbf{0.6055}} & \mcc{\textbf{0.6260}} \\
\cmidrule(l){2-16}
\multirow{5}{*}{\rotatebox{90}{\shortstack{EgoExo-\\Fitness}}}
 & Qwen3-VL-Instruct\vphantom{$^\dagger$}
   & \mcc{\textbf{0.5733}} & \mcc{0.5643} & \mcc{0.5498} & \mcc{0.5363} & \mcc{0.5719} & \mcc{\textbf{0.5776}} & \mcc{0.5220} \\
 & Qwen3-VL-Thinking\vphantom{$^\dagger$}
   & \mcc{\textbf{0.5650}} & \mcc{\textbf{0.5633}} & \mcc{0.5556} & \mcc{0.5612} & \mcc{0.5507} & \mcc{0.5491} & \mcc{0.5007} \\
 & InternVL3.5\vphantom{$^\dagger$}
   & \mcc{0.5429} & \mcc{0.5452} & \mcc{0.5323} & \mcc{0.5362} & \mcc{\textbf{0.5657}} & \mcc{\textbf{0.5586}} & \mcc{--$^*$} \\
 & InternVL3.5$^\dagger$
   & \mcc{0.5461} & \mcc{0.5427} & \mcc{0.5378} & \mcc{0.5441} & \mcc{\textbf{0.5516}} & \mcc{\textbf{0.5468}} & \mcc{--$^*$} \\
 & Gemini 3.1 Pro\vphantom{$^\dagger$}
   & \mcc{0.5167} & \mcc{0.5314} & \mcc{\textbf{0.5442}} & \mcc{\textbf{0.5497}} & \mcc{0.5304} & \mcc{0.5090} & \mcc{0.4603} \\
\cmidrule(l){2-16}
\multirow{5}{*}{\rotatebox{90}{\shortstack{Fitness-\\AQA}}}
 & Qwen3-VL-Instruct\vphantom{$^\dagger$}
   & \mcc{\textbf{0.5557}} & \mcc{0.5400} & \mcc{0.5300} & \mcc{0.5293} & \mcc{0.5482} & \mcc{\textbf{0.5545}} & \mcc{0.5379} \\
 & Qwen3-VL-Thinking\vphantom{$^\dagger$}
   & \mcc{0.5224} & \mcc{\textbf{0.5351}} & \mcc{\textbf{0.5375}} & \mcc{0.5060} & \mcc{0.5215} & \mcc{0.5162} & \mcc{0.5336} \\
 & InternVL3.5\vphantom{$^\dagger$}
   & \mcc{0.5296} & \mcc{0.5077} & \mcc{0.4970} & \mcc{0.5085} & \mcc{\textbf{0.5507}} & \mcc{\textbf{0.5335}} & \mcc{--$^*$} \\
 & InternVL3.5$^\dagger$
   & \mcc{\textbf{0.5131}} & \mcc{0.4966} & \mcc{0.5046} & \mcc{0.4991} & \mcc{0.5041} & \mcc{\textbf{0.5066}} & \mcc{--$^*$} \\
 & Gemini 3.1 Pro\vphantom{$^\dagger$}
   & \mcc{0.5596} & \mcc{0.5623} & \mcc{0.5432} & \mcc{0.5583} & \mcc{0.5538} & \mcc{\textbf{0.5700}} & \mcc{\textbf{0.5744}} \\
\midrule
\textit{Regression} &
  & $\rho\uparrow$ & $\text{R-}\ell_2\downarrow$
  & $\rho\uparrow$ & $\text{R-}\ell_2\downarrow$
  & $\rho\uparrow$ & $\text{R-}\ell_2\downarrow$
  & $\rho\uparrow$ & $\text{R-}\ell_2\downarrow$
  & $\rho\uparrow$ & $\text{R-}\ell_2\downarrow$
  & $\rho\uparrow$ & $\text{R-}\ell_2\downarrow$
  & $\rho\uparrow$ & $\text{R-}\ell_2\downarrow$ \\[2pt]
\multirow{5}{*}{\rotatebox{90}{FineFS}}
 & Qwen3-VL-Instruct\vphantom{$^\dagger$}
   & 0.1998 & 0.2590 & \textbf{0.2228} & 0.2571 & 0.1416 & 0.3788 & 0.2061 & 0.3785 & 0.1600 & 0.2789 & 0.2007 & \textbf{0.2562} & \textbf{0.3120} & \textbf{0.1716} \\
 & Qwen3-VL-Thinking\vphantom{$^\dagger$}
   & \textbf{0.2806} & 0.2673 & 0.2202 & 0.2722 & 0.1849 & \textbf{0.2581} & 0.1685 & 0.3313 & 0.1841 & 0.2809 & 0.2677 & 0.2744 & \textbf{0.3020} & \textbf{0.1625} \\
 & InternVL3.5\vphantom{$^\dagger$}
   & 0.2042 & \textbf{0.2540} & \textbf{0.2433} & 0.2721 & 0.1500 & 0.3333 & 0.2235 & 0.3063 & \textbf{0.2703} & 0.2596 & 0.2120 & \textbf{0.2567} & --$^*$ & --$^*$ \\
 & InternVL3.5$^\dagger$
   & \textbf{0.2797} & 0.2372 & 0.2750 & \textbf{0.2325} & 0.2382 & 0.2684 & 0.2651 & 0.2486 & \textbf{0.2931} & 0.2429 & 0.2723 & \textbf{0.2197} & --$^*$ & --$^*$ \\
 & Gemini 3.1 Pro\vphantom{$^\dagger$}
   & \textbf{0.3690} & 0.3984 & 0.2651 & 0.3079 & 0.2739 & 0.3097 & 0.3164 & 0.3230 & 0.2931 & 0.3151 & 0.2891 & \textbf{0.3047} & \textbf{0.3184} & \textbf{0.2772} \\
\cmidrule(l){2-16}
\multirow{5}{*}{\rotatebox{90}{MTL-AQA}}
 & Qwen3-VL-Instruct\vphantom{$^\dagger$}
   & 0.2319 & 0.2026 & 0.2484 & 0.1813 & 0.1682 & 0.3062 & \textbf{0.2679} & 0.2834 & \textbf{0.3239} & 0.2119 & \textbf{0.3505} & 0.1996 & 0.1426 & \textbf{0.1780} \\
 & Qwen3-VL-Thinking\vphantom{$^\dagger$}
   & 0.1367 & \textbf{0.2665} & \textbf{0.2308} & 0.2750 & 0.1264 & 0.2910 & 0.1037 & 0.2849 & 0.0865 & 0.2754 & 0.0982 & 0.2822 & \textbf{0.2182} & \textbf{0.2026} \\
 & InternVL3.5\vphantom{$^\dagger$}
   & \textbf{0.2636} & 0.1847 & 0.1410 & 0.1783 & \textbf{0.3325} & 0.2364 & 0.1239 & 0.2595 & 0.2561 & \textbf{0.1730} & 0.2314 & \textbf{0.1762} & 0.1505 & 0.1860 \\
 & InternVL3.5$^\dagger$
   & 0.1974 & 0.2487 & 0.2179 & 0.2509 & \textbf{0.2711} & 0.2288 & 0.2535 & 0.2269 & \textbf{0.3406} & \textbf{0.2081} & 0.2548 & 0.2175 & 0.0238 & \textbf{0.1836} \\
 & Gemini 3.1 Pro\vphantom{$^\dagger$}
   & 0.0576 & 0.3060 & 0.0788 & 0.3145 & 0.0579 & \textbf{0.3016} & 0.0379 & \textbf{0.2528} & \textbf{0.1046} & 0.3151 & -0.0162 & 0.3426 & \textbf{0.2057} & 0.3019 \\
\bottomrule
\multicolumn{16}{@{}l}{$^\dagger$ thinking mode. $^*$ could not be evaluated due to context window limitations.}
\end{tabular*}
\caption{\textbf{Results across prompt engineering strategies.} \textbf{Classification }tasks (LLM-FMS, EgoExo-Fitness, Fitness-AQA) are evaluated with balanced accuracy (\%). \textbf{Regression} tasks (FineFS, MTL-AQA) are evaluated with Spearman correlation ($\rho$) and R-$\ell_2$. Two best results per row for each metric are shown in \textbf{bold}.}
\label{tab:prompt_results}
\end{table*}

\paragraph{Classification results.} Results in Table \ref{tab:prompt_results} show that prompts designed to encourage deeper visual analysis, namely \textit{Visual Grounding}, \textit{Two-Step}, and \textit{Structured Reasoning}, yield little to no performance improvement. Adding exercise knowledge through positive or negative guidelines leads to marginal and inconsistent gains. In-context learning stands out as the most reliable technique when applied to the image dataset (LLM-FMS), delivering consistent low-to-moderate improvements across models; however, it fails to generalise to video datasets. A possible explanation is that models struggle with the significant extension of context length introduced by the few-shot examples.

\paragraph{Regression results.} In regression tasks, \textit{Visual Grounding}, \textit{Two-Step}, and \textit{Structured Reasoning} prompts again yield inconsistent results. Positive guidelines, however, often produce a considerably stronger effect than in classification tasks. This is likely because these tasks require holistic evaluations of complex diving and figure skating performances, rather than the more focused analyses of the classification tasks, a distinction that increases the value of pre-defined, objective criteria. In-context learning  also has a notably higher impact in these tasks, particularly on the FineFS dataset, where it consistently achieves substantial improvements in both rank correlation and R-$\ell_2$ norm. Two factors can explain this: the inherent complexity of the activity domains, where intricate movements are difficult to assess without reference examples, and the abstract nature of score regression, where even two examples can meaningfully anchor score calibration.

\paragraph{Summary.} Overall, these results indicate that standard prompting strategies offer limited benefit. Guideline-based prompting occasionally improves performance, but remains unreliable. In-context learning, by contrast, can yield meaningful gains particularly for regression tasks and image data.
\begin{figure*}[t]
    \centering
    \vspace{0.2cm}
    \includegraphics[width=\textwidth]{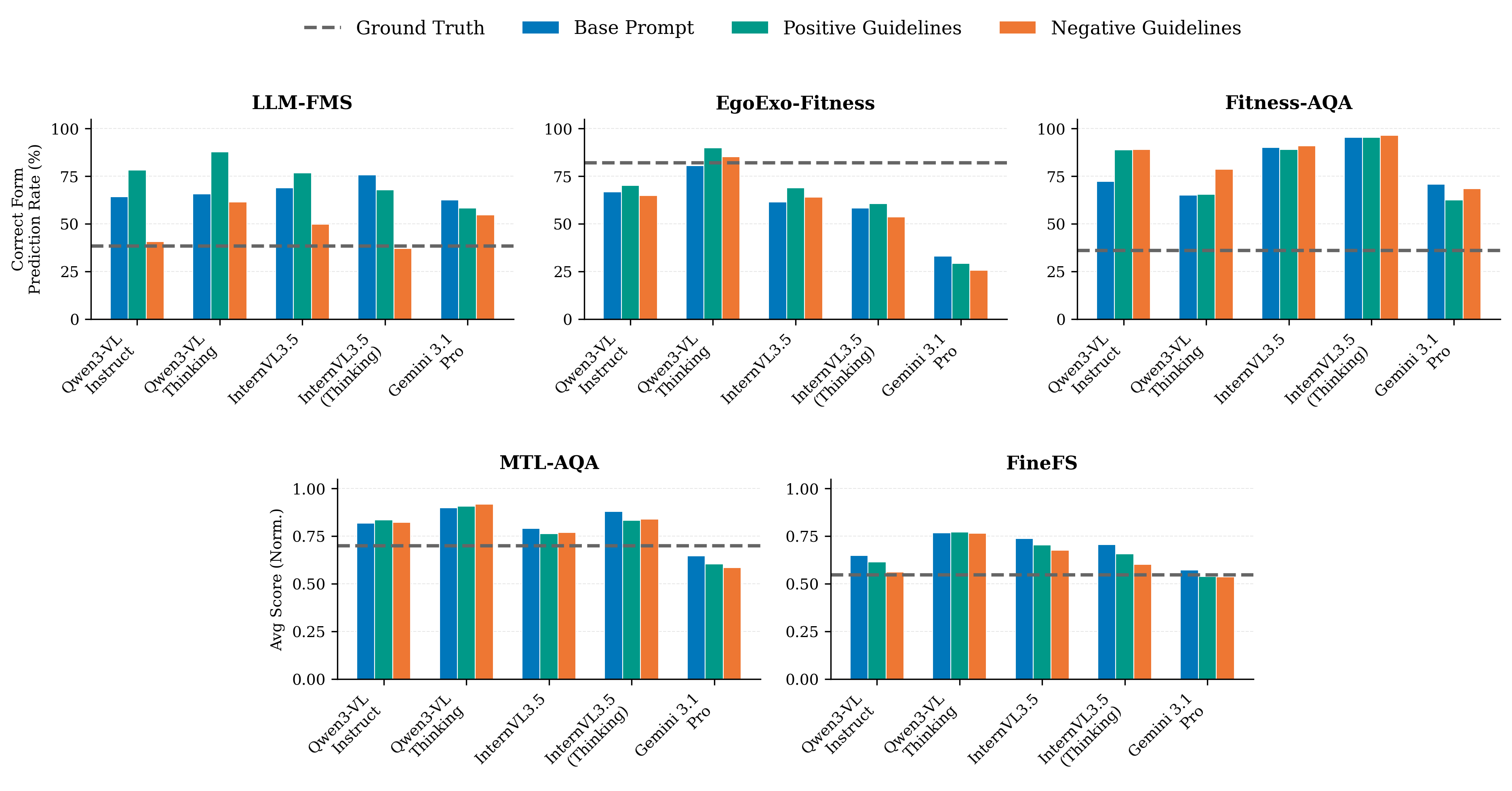}
    \caption{\textbf{Prompting effects on model predictions.} For \textbf{classification tasks (top)}, each subplot shows the percentage of predictions matching the correct execution form. For \textbf{regression tasks (bottom)}, each subplot shows the mean predicted score normalised to 0–1. In both cases, results are shown across all models and three prompting strategies: \textbf{base prompt (blue), positive guidelines (green), and negative guidelines (orange)}, with dashed lines indicating ground truth values.}
    \label{fig:prompt_bias}
    \vspace{0.2cm}
\end{figure*}
\subsection{Prompt Bias}
\label{sec:prmpt_bias}
While prompt engineering can guide model behaviour in beneficial ways, it may also introduce unintended biases. To quantify the influence of language priors on VLM-based AQA, we compare model predictions under two prompt variants: \textit{positive guidelines}, which describe a correct execution, and \textit{negative guidelines}, which describe common errors. The two variants differ only in the polarity of their phrasing, conveying identical semantic content. For example, a positive guideline might state ``\textit{The trunk should be parallel to the calf}", while its negative counterpart would be ``\textit{The trunk not being parallel to the calf}". Any systematic shift in predictions between the two variants can therefore be attributed to language priors rather than to differences in the information provided.

For classification tasks, we measure how often models predict correct execution under each variant. For EgoExo-Fitness and Fitness-AQA, the answer associated with correct execution is inferred directly from the existing annotations. For LLM-FMS, we prompted Claude Sonnet 4.6 \cite{anthropic2026sonnet46systemcard} to determine, for each question, which answer option corresponds to correct execution of the exercise. For the regression tasks, we measure the shift in the average predicted score between the two prompt variants.

\paragraph{Correctness bias.} An important initial observation is that, across most datasets, models exhibit a strong tendency to predict that exercises are performed correctly. As shown in Figure~\ref{fig:prompt_bias}, the proportion of predictions corresponding to correct execution substantially exceeds the ground truth rate in both LLM-FMS and Fitness-AQA, and average predicted scores are similarly inflated in the regression tasks. A possible explanation is that VLMs rely on prior knowledge of how an exercise should look, biasing them towards predicting correct execution regardless of the visual content.

\paragraph{Sensitivity to guidelines.} Figure~\ref{fig:prompt_bias} further reveals systematic variations in prediction rates across the three prompt variants. For LLM-FMS and EgoExo-Fitness, models are more likely to predict correct execution under positive guidelines and incorrect execution under negative guidelines. For Fitness-AQA, this pattern is consistently inverted, which can be attributed to the error-detection nature of the task: explicit descriptions of poor execution may raise the model's threshold for what it considers an erroneous performance. Taken together, the consistent shifts across all classification tasks strongly suggest that models are heavily influenced by prompt phrasing, exploiting linguistic cues rather than relying on visual grounding alone.

\paragraph{Regression is less sensitive to linguistic cues.} These effects are markedly less pronounced in regression tasks, where the differences in predicted scores between prompt variants are often minimal. This indicates that prompt phrasing exerts less influence on score regression than on classification tasks. A plausible explanation is that linguistic cues more readily bias language-based tasks requiring the interpretation of questions, textual instructions, or error descriptions, whereas generating a continuous numerical prediction may be less susceptible to such shortcuts.

\subsection{Contrastive Tasks}
One way to mitigate these biases is to reformulate tasks as contrastive comparisons. In this setting, prior knowledge and linguistic cues provide little basis for distinguishing which of two samples demonstrates better execution. We reformulate all datasets and tasks into this framework.

\begin{itemize}
\item \textbf{LLM-FMS:} We rephrase all questions in a contrastive format. For instance,  ``\textit{What is the height of the hip relative to the knee on the vertical axis?}'' with options \textit{Higher}, \textit{Equal}, and \textit{Lower} is reformulated as ``\textit{In which image is the hip lower relative to the knee?}''. We ensure the two samples always differ in their answers to the original question, making the contrastive task unambiguous. To further simplify the task, we exclude boundary options such as \textit{Equal}, which represent unclear edge cases.

\item \textbf{EgoExo-Fitness:} For each video and technical guideline, we randomly select a second video whose ground-truth label for that keypoint is opposite to that of the first. The model is then prompted to identify which of the two videos better adheres to the given guideline.

\item \textbf{Fitness-AQA:} For each video in which no execution errors occur, we randomly pair it with a video containing at least one error, and vice versa. The model is prompted to select the better execution, with the error-free sample treated as the ground truth.

\item \textbf{FineFS:} Each sample is paired with a second video depicting the same type of figure skating element. The model is asked to identify the better execution, using the sample with the higher GOE as ground truth.

\item \textbf{MTL-AQA:} Each dive sample is paired with a second dive, and the model is prompted to select the better execution, using the higher-scored sample as ground truth.
\end{itemize}

\begin{table}[t]
\centering
\renewcommand{\arraystretch}{0.85}
\setlength{\tabcolsep}{0pt}
\scriptsize
\begin{tabular*}{\columnwidth}{@{} l @{\extracolsep{\fill}} c c c c c @{\hspace{4pt}}}
\toprule
\textbf{Model} & \textbf{LLM-FMS} & \textbf{\shortstack{EgoExo-\\Fitness}} & \textbf{\shortstack{Fitness-\\AQA}} & \textbf{FineFS} & \textbf{MTL-AQA} \\
\midrule
Random guess\vphantom{$^\dagger$}           & 50.00 & 50.00 & 50.00 & 50.00 & 50.00 \\
\midrule
Qwen3-VL-Instruct\vphantom{$^\dagger$}      & 68.97 & \textbf{54.61} & 50.60 & 56.40 & 60.06 \\
Qwen3-VL-Thinking\vphantom{$^\dagger$}      & 60.87 & 54.24 & 51.80 & 57.40 & 55.52 \\
InternVL3.5\vphantom{$^\dagger$}            & 62.55 & --$^*$ & 47.17 & --$^*$ & 52.41 \\
InternVL3.5$^\dagger$                       & 59.98 & --$^*$ & 49.74 & --$^*$ & 53.26 \\
Gemini 3.1 Pro\vphantom{$^\dagger$}         & \textbf{80.12} & 51.20 & \textbf{52.49} & \textbf{58.91} & \textbf{61.76} \\
\bottomrule
\multicolumn{6}{@{}l}{$^\dagger$ thinking mode. $^*$ could not be evaluated due to context window limitations.}
\end{tabular*}
\caption{Accuracy (\%) on contrastive tasks, with best result per column shown in \textbf{bold}.}
\label{tab:contrastive}
\end{table}

\begin{figure*}[t]
    \centering
    \includegraphics[width=\textwidth]{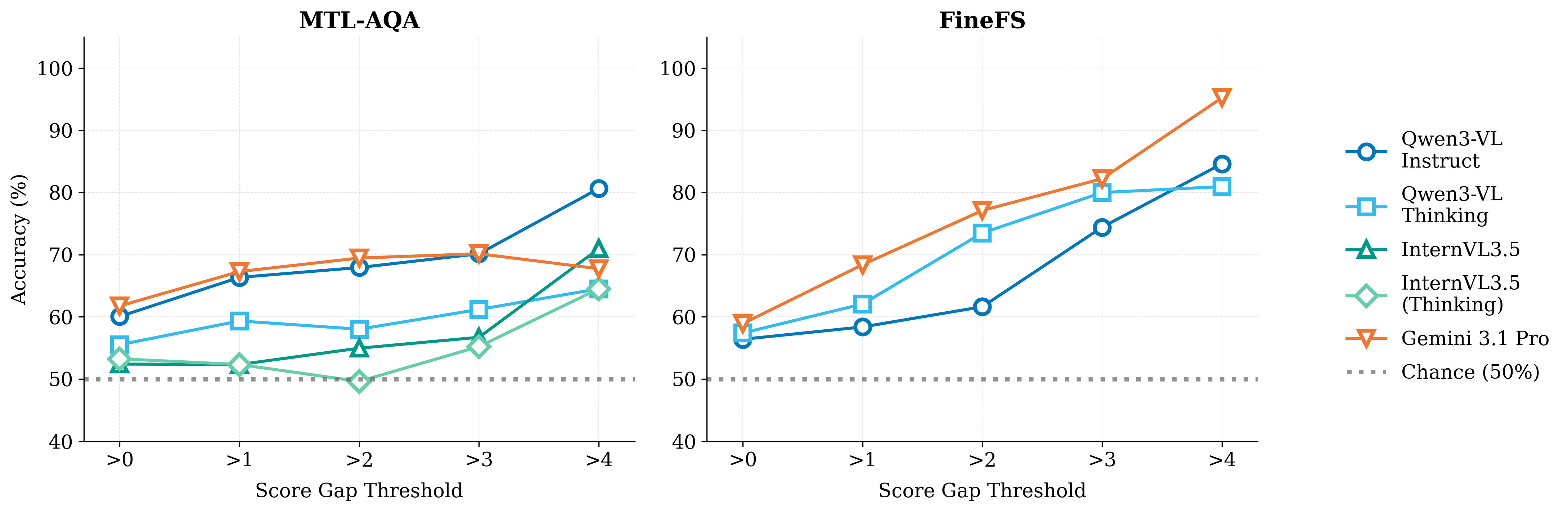}
    \caption{\textbf{Accuracy (\%) on contrastive tasks for MTL-AQA and FineFS as a function of the score gap between samples.} Larger gaps generally correspond to easier comparisons, as the quality difference between samples becomes more pronounced.}
    \label{fig:contrastive_accuracy_per_score_gap}
\end{figure*}

\paragraph{Results.} Consistent with non-contrastive tasks, results (Table \ref{tab:contrastive}) show that most models perform only marginally above chance (with the notable exceptions of Gemini 3.1 Pro and Qwen3-VL-Instruct on LLM-FMS). This suggests that the observed poor performance is not a direct consequence of the identified biases, but rather reflects inherently limited AQA capabilities.

Nevertheless, Figure \ref{fig:contrastive_accuracy_per_score_gap} shows a clear positive relationship between model accuracy and the score gap between compared samples. When the gap is pronounced, models achieve considerably high accuracies, indicating that they possess some AQA capability, but it degrades substantially as comparisons become less obvious.

\subsection{Qualitative Analysis of Reasoning Traces}
Manual inspection of reasoning traces for Qwen3-VL-Thinking and InternVL3.5 
(in thinking mode) across datasets reveals that these models consistently 
mention prior knowledge of the exercise throughout the reasoning:

\begin{tcolorbox}[
    colback=gray!8,
    colframe=gray!40,
    arc=3pt,
    boxsep=2pt,
    left=6pt, right=6pt, top=5pt, bottom=5pt,
    fontupper=\small\itshape
]
``In a standard push-up, the elbows are bent, and the torso is close to the ground.'' \\[6pt]
``The person is in a squat, so the trunk is likely parallel.'' \\[6pt]
``Overhead press is usually standing.`` \\[6pt]
``Higher jumps usually get better GOE.'' \\[6pt] 
``Considering the Olympics, the execution is likely high.``
\end{tcolorbox}

While some references may reflect legitimate reasoning that aids the model 
in its task, they may also indicate a tendency to ground answers in prior 
knowledge rather than visual analysis. This hypothesis aligns with the 
observed correctness bias discussed in Section \ref{sec:prmpt_bias}.

\section{Related Work}
\label{sec:related_work}

\paragraph{AQA as score regression.}
AQA has predominantly been framed as a regression problem, with approaches evolving from global 3D CNN features \cite{parmar2017learning, carreira2017quo} to segment-aware and actor-centric representations \cite{xiang2018s3d, wang2021tsa}, and more recently transformer-based architectures for long-range temporal modelling \cite{xu2022likert, fang2023end}. Despite strong performance, these methods offer no actionable feedback on how to improve execution.

\paragraph{AQA with text generation.}
Text generation in AQA began with commentary-paired datasets \cite{parmar2019what} and has since advanced toward more structured, feedback-centric approaches that generate narrative evaluations \cite{zhang2024narrative, li2025techcoach}. However, these methods lack the expressivity of LLMs and remain tightly constrained to their training formats, limiting generalisability.

\paragraph{VLM-based AQA.}
VLMs have seen limited exploration in AQA. Early work demonstrated their viability on a small domain \cite{noworolnik2025assessing}, while \citet{wu2025hieroaction} extended this through fine-tuning with structured chain-of-thought reasoning and hierarchical reinforcement learning, yielding improvements over zero-shot baselines but falling short of specialised models. Despite this progress, key open questions remain around skeletal feature integration, prompt engineering, and model biases in VLM-based AQA.
\section{Discussion}
\label{sec:discussion}

\paragraph{Performance overview.} Baseline results demonstrate that off-the-shelf VLMs achieve only marginal improvements over random chance across a broad range of datasets and tasks. Incorporating skeleton information can boost performance in some cases, but the effect is small. Similarly, prompt engineering yields limited gains with no single strategy proving reliably superior. In-context learning is the only technique to produce meaningful improvements on certain tasks, but its applicability is constrained by the substantial token overhead introduced by additional video examples.

\paragraph{Systematic biases.} Prediction distributions reveal two systematic biases in VLMs. The first is a strong tendency to predict that exercises are performed correctly, regardless of visual content, likely stemming from the models' reliance on prior knowledge of how exercises should look. This is corroborated by the analysis of reasoning traces, where references to prior knowledge, such as \textit{"The person is in a squat, so the trunk is likely parallel"}, appear frequently. The second is a susceptibility to linguistic cues: minor rephrasing of guidelines produces consistent shifts in predictions, despite no change in semantic content. This aligns with prior work documenting VLMs' tendency to over-rely on the language modality \cite{sim2025vlms}, and represents a key limitation for the application of general-purpose VLMs to AQA.

\paragraph{Contrastive reformulation.} A contrastive reformulation mitigates both biases, as neither prior knowledge nor prompt phrasing provides a useful signal for determining which of two videos (or images) shows superior execution. Nevertheless, performance remains poor, indicating that while these biases are real, they are not the primary driver of the models' underperformance.

\paragraph{Takeaways.} These findings indicate that general-purpose VLMs fundamentally struggle with AQA tasks, an observation that holds consistently across optimisation strategies and task reformulations. Thus, substantial progress in model capabilities or training approaches will be required before VLMs become viable for AQA applications.

\paragraph{Limitations.}
This study focuses on off-the-shelf VLMs; whether fine-tuning on domain-specific data could overcome the observed biases and performance gaps remains an open question. Furthermore, the evaluation is limited to three model families, and the observed patterns may not generalise to other architectures or training regimes.
\section{Conclusion}
\label{sec:conclusion}

The application of VLMs to AQA holds considerable promise, thanks to their prompting-based flexibility, conversational capabilities, and potential for explainability. However, the comprehensive assessment undertaken makes it clear that even state-of-the-art models remain unreliable for real-world deployment. Preprocessing techniques such as cropping and prompting strategies such as in-context learning can yield marginal improvements in specific settings, but no strategy proves consistently effective across models or tasks. Moreover, these models exhibit two systematic biases: an over-reliance on prior knowledge and an excessive sensitivity to linguistic framing. Still, in contrastive settings designed to mitigate these biases, models continue to exhibit poor performance, suggesting that the underlying limitations run deeper than the identified biases alone. 

Substantial advancements in model capabilities will be required in future work before VLMs can be reliably applied to AQA. In this light, this study provides a reference benchmark for assessing and guiding further developments in VLM-based AQA.

\noindent \textbf{Acknowledgements.}
\small This research was funded by Sword Health and further supported by Recovery and Resilience Plan and Fundação para a Ciência e a Tecnologia (FCT) through the ATE project (02/C05-i01.02/2022 under agreement PC644914747-00000023) and INESC-ID pluriannual (UID/PRR/50021/2025 and UID/50021/2025).
{
    \small
    \bibliographystyle{ieeenat_fullname}
    \bibliography{main}
}


\clearpage
\onecolumn

\begin{center}
  {\Large\bfseries Can Vision Language Models Judge Action Quality? An Empirical Evaluation}\\[0.6em]
  {\large Appendix}
\end{center}

\appendix
\setcounter{figure}{0}
\renewcommand{\thefigure}{A.\arabic{figure}}

\section{Dataset Preprocessing Examples}
\label{sec:appendix_datasets}

We show representative frames from each dataset under four
preprocessing methods: original RGB frame, cropped
frame, skeleton overlay on the RGB image, and skeleton render.

\subsection{LLM-FMS}
\label{sec:appendix_llmfms}

\newsavebox{\imgA}\newsavebox{\imgB}\newsavebox{\imgC}\newsavebox{\imgD}
\begin{figure}[h]
  \centering
  \sbox{\imgA}{\includegraphics[height=3cm]{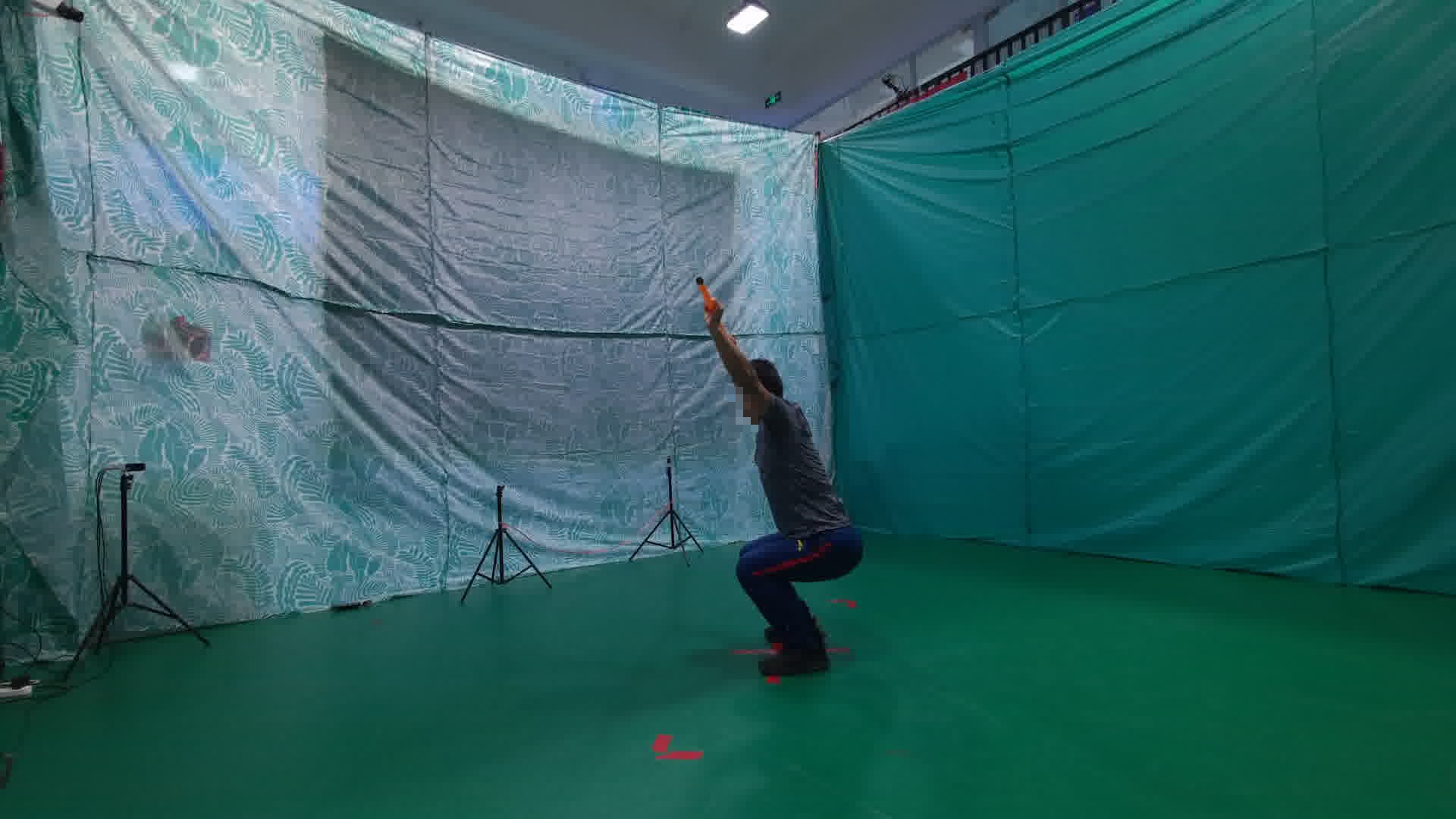}}%
  \sbox{\imgB}{\includegraphics[height=3cm]{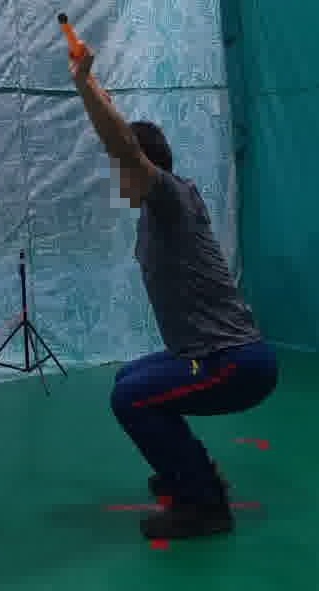}}%
  \sbox{\imgC}{\includegraphics[height=3cm]{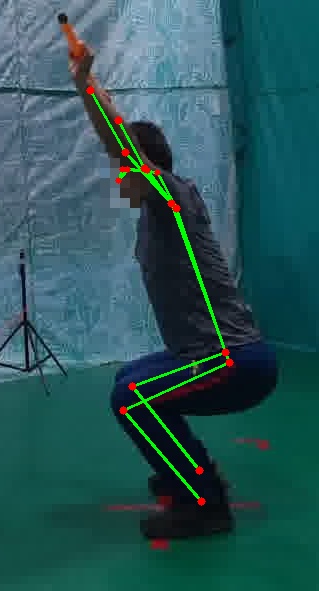}}%
  \sbox{\imgD}{\fbox{\includegraphics[height=3cm]{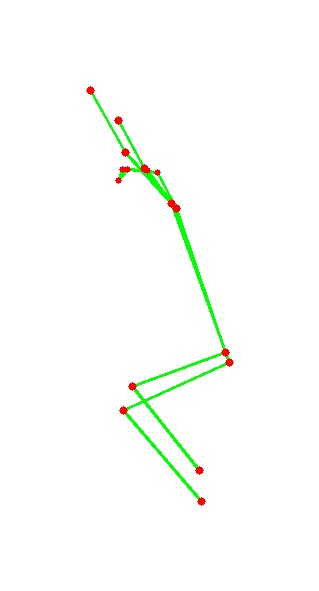}}}%
  \usebox{\imgA}\hfill\usebox{\imgB}\hfill\usebox{\imgC}\hfill\usebox{\imgD}\\[4pt]
  \makebox[\wd\imgA]{\small Original}\hfill
  \makebox[\wd\imgB]{\small Cropped}\hfill
  \makebox[\wd\imgC]{\small Skeleton Overlay}\hfill
  \makebox[\wd\imgD]{\small Skeleton Render}
  \caption{LLM-FMS Preprocessing Examples.}
  \label{fig:llmfms_samples}
\end{figure}

\subsection{EgoExo-Fitness}
\label{sec:appendix_egoexo}

\begin{figure}[h]
  \centering
  \sbox{\imgA}{\includegraphics[height=3cm]{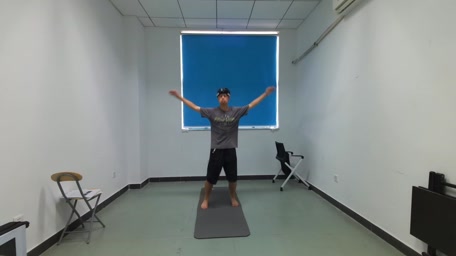}}%
  \sbox{\imgB}{\includegraphics[height=3cm]{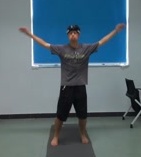}}%
  \sbox{\imgC}{\includegraphics[height=3cm]{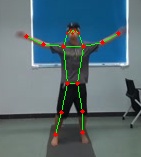}}%
  \sbox{\imgD}{\fbox{\includegraphics[height=3cm]{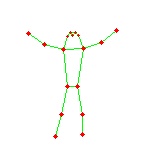}}}%
  \usebox{\imgA}\hfill\usebox{\imgB}\hfill\usebox{\imgC}\hfill\usebox{\imgD}\\[4pt]
  \makebox[\wd\imgA]{\small Original}\hfill
  \makebox[\wd\imgB]{\small Cropped}\hfill
  \makebox[\wd\imgC]{\small Skeleton Overlay}\hfill
  \makebox[\wd\imgD]{\small Skeleton Render}
  \caption{EgoExo-Fitness Preprocessing Examples.}
  \label{fig:egoexo_samples}
\end{figure}

\subsection{Fitness-AQA}
\label{sec:appendix_fitness}

\begin{figure}[h]
  \centering
  \sbox{\imgA}{\includegraphics[height=3cm]{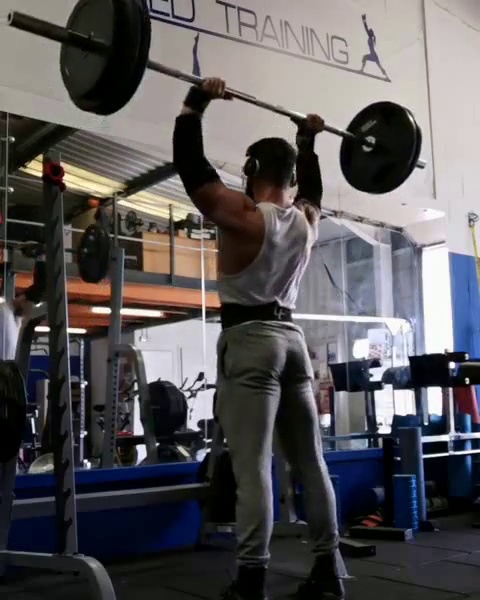}}%
  \sbox{\imgB}{\includegraphics[height=3cm]{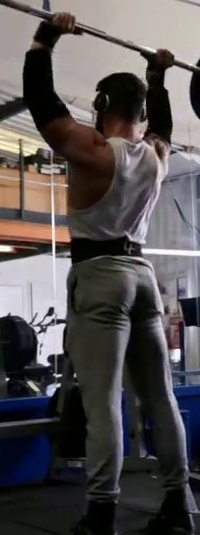}}%
  \sbox{\imgC}{\includegraphics[height=3cm]{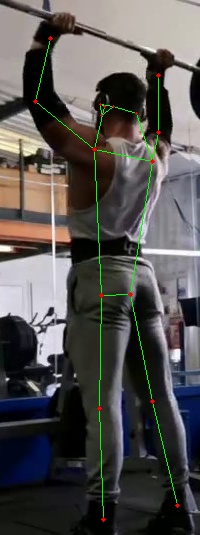}}%
  \sbox{\imgD}{\fbox{\includegraphics[height=3cm]{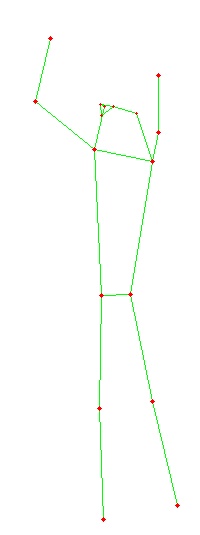}}}%
  \usebox{\imgA}\hfill\usebox{\imgB}\hfill\usebox{\imgC}\hfill\usebox{\imgD}\\[4pt]
  \makebox[\wd\imgA]{\small Original}\hfill
  \makebox[\wd\imgB]{\small Cropped}\hfill
  \makebox[\wd\imgC]{\small Skeleton Overlay}\hfill
  \makebox[\wd\imgD]{\small Skeleton Render}
  \caption{Fitness-AQA Preprocessing Examples.}
  \label{fig:fitness_samples}
\end{figure}

\clearpage
\subsection{FineFS}
\label{sec:appendix_finefs}

\refstepcounter{figure}
\begin{center}
  \includegraphics[width=\linewidth,height=0.24\textheight,keepaspectratio]{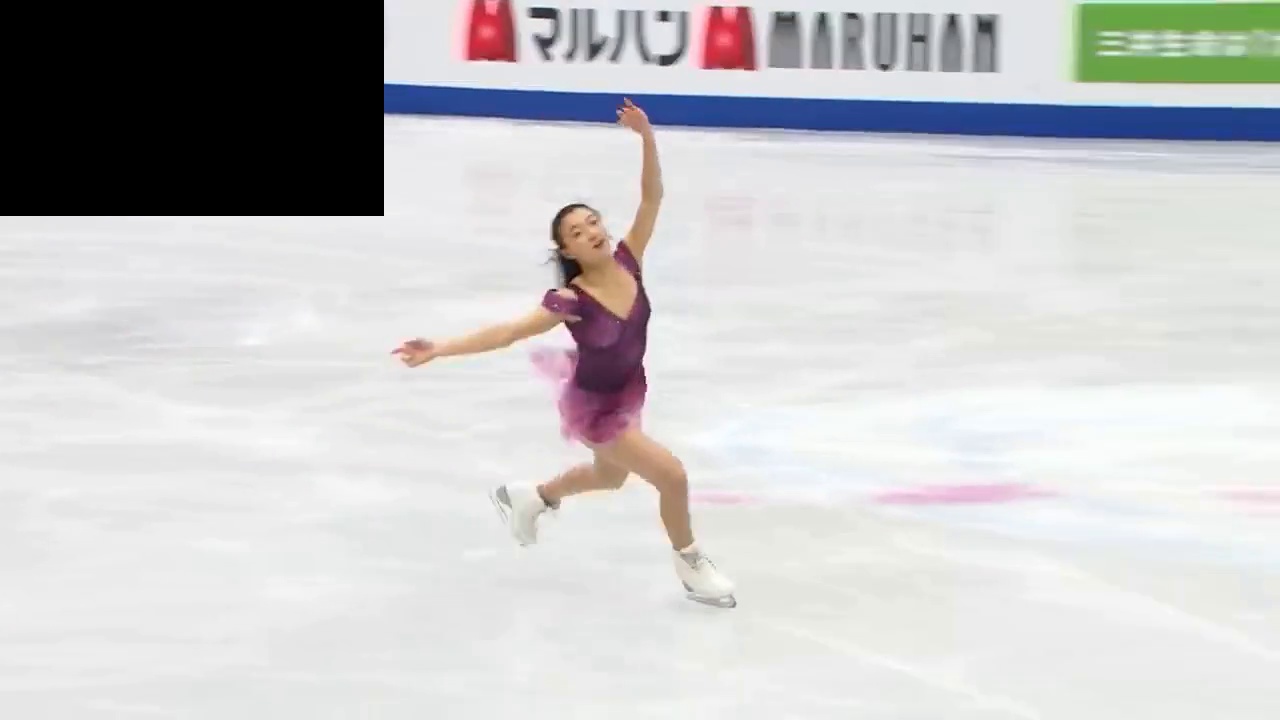}\\[4pt]
  {\small Original}

  \vspace{1em}

  \includegraphics[width=\linewidth,height=0.24\textheight,keepaspectratio]{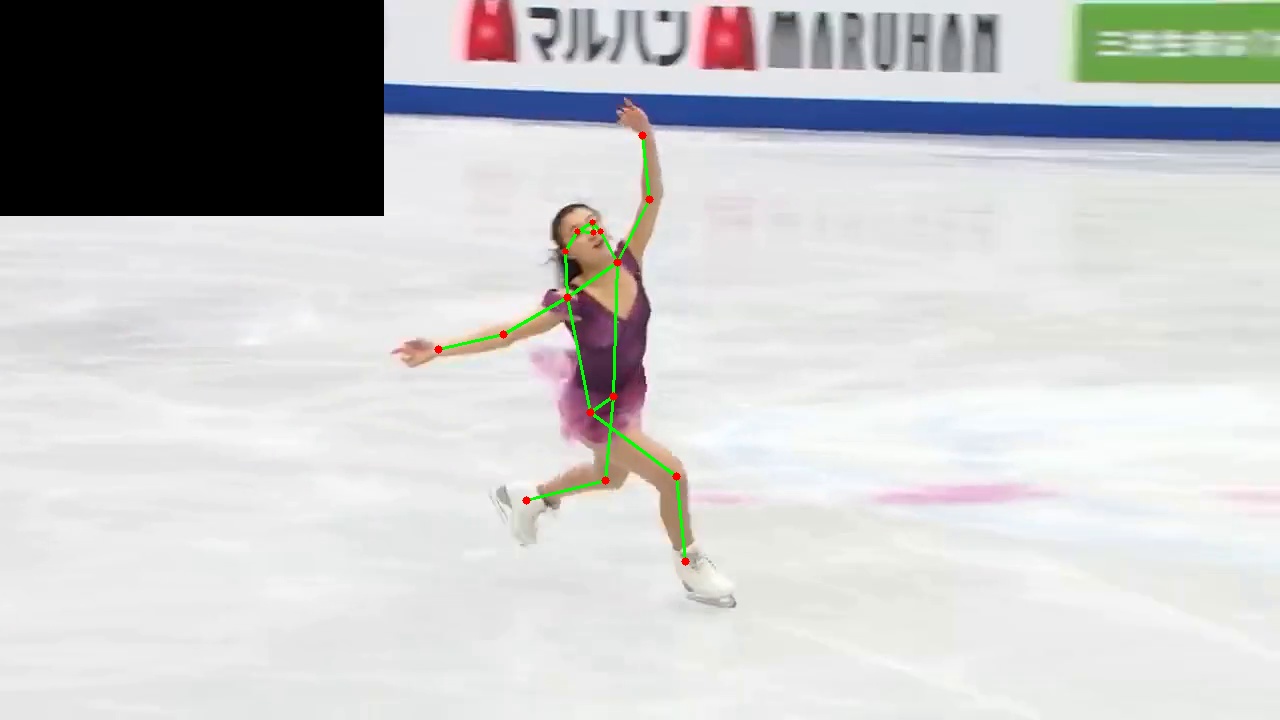}\\[4pt]
  {\small Skeleton Overlay}

  \vspace{1em}

  \fbox{\includegraphics[width=\linewidth,height=0.24\textheight,keepaspectratio]{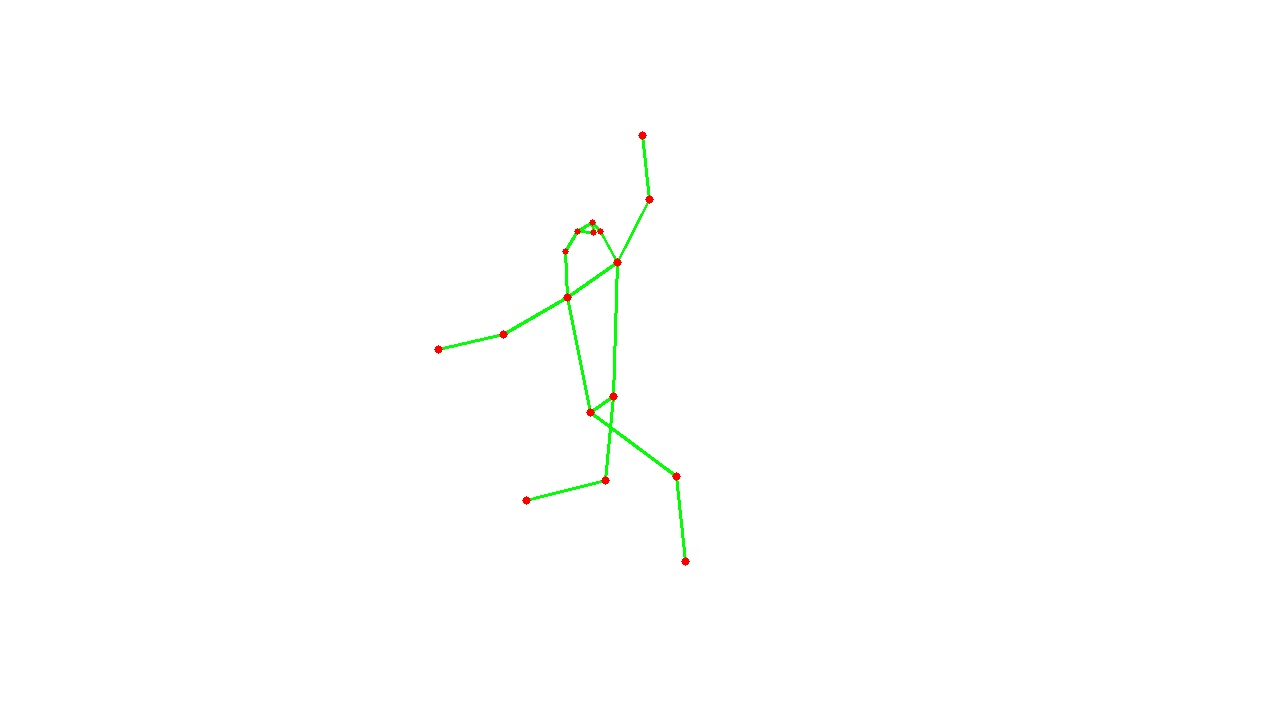}}\\[4pt]
  {\small Skeleton Render}

  \vspace{0.5em}
  {\small Figure \thefigure: FineFS Preprocessing Examples.}
  \label{fig:finefs_samples}
\end{center}

\clearpage
\subsection{MTL-AQA}
\label{sec:appendix_mtlaqa}

\refstepcounter{figure}
\begin{center}
  \includegraphics[width=\linewidth,height=0.24\textheight,keepaspectratio]{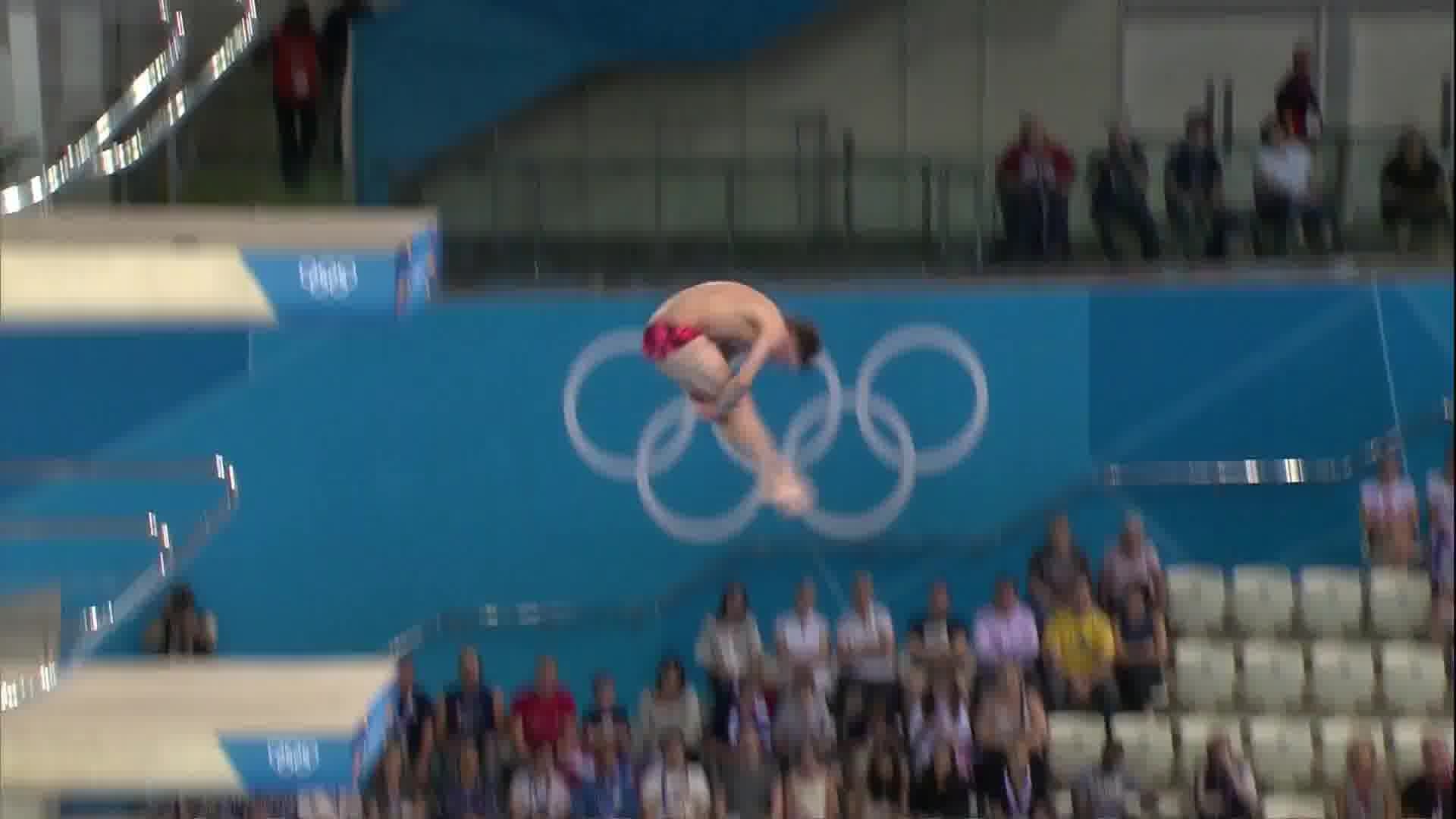}\\[4pt]
  {\small Original}

  \vspace{1em}

  \includegraphics[width=\linewidth,height=0.24\textheight,keepaspectratio]{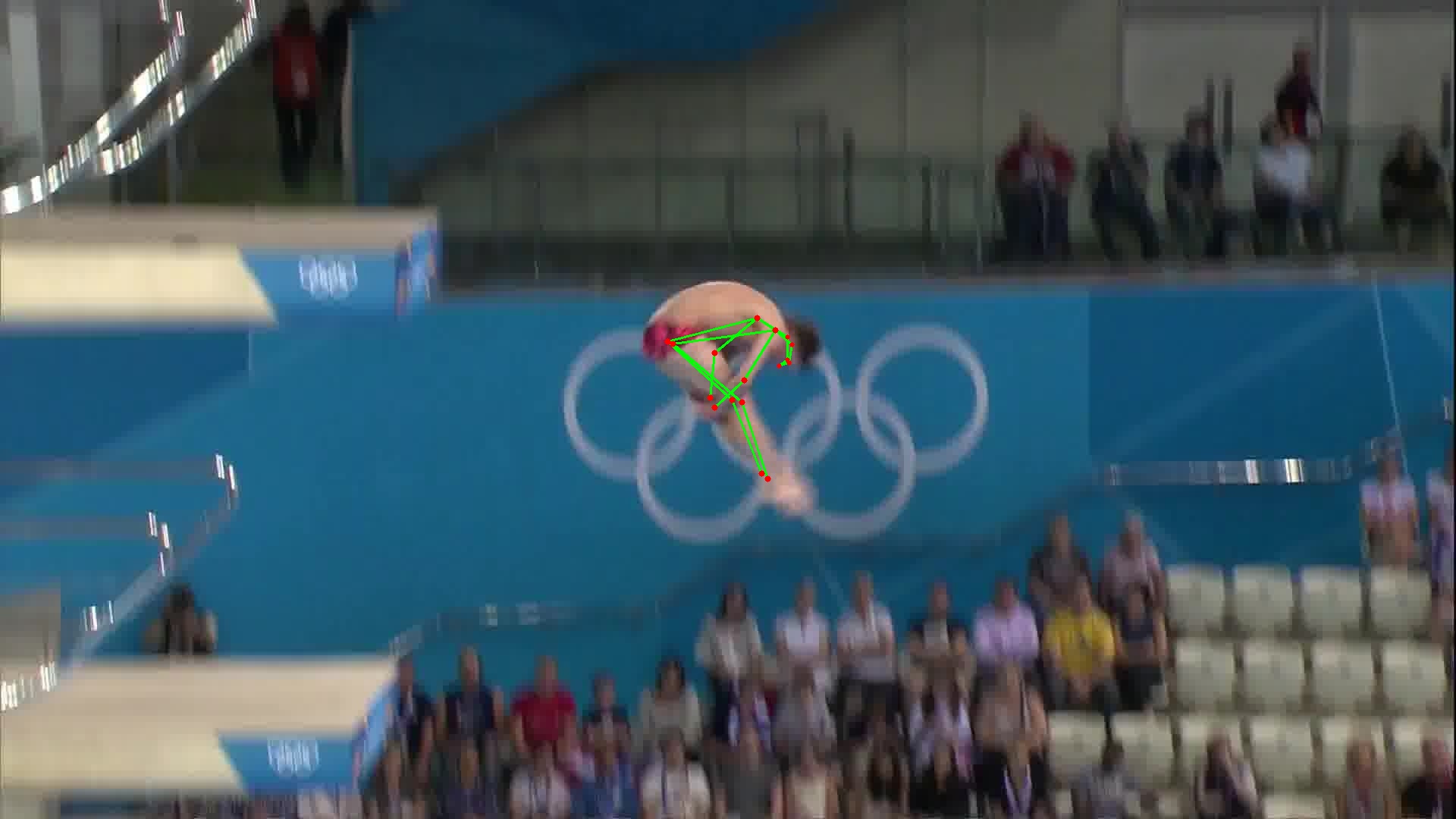}\\[4pt]
  {\small Skeleton Overlay}

  \vspace{1em}

  \fbox{\includegraphics[width=\linewidth,height=0.24\textheight,keepaspectratio]{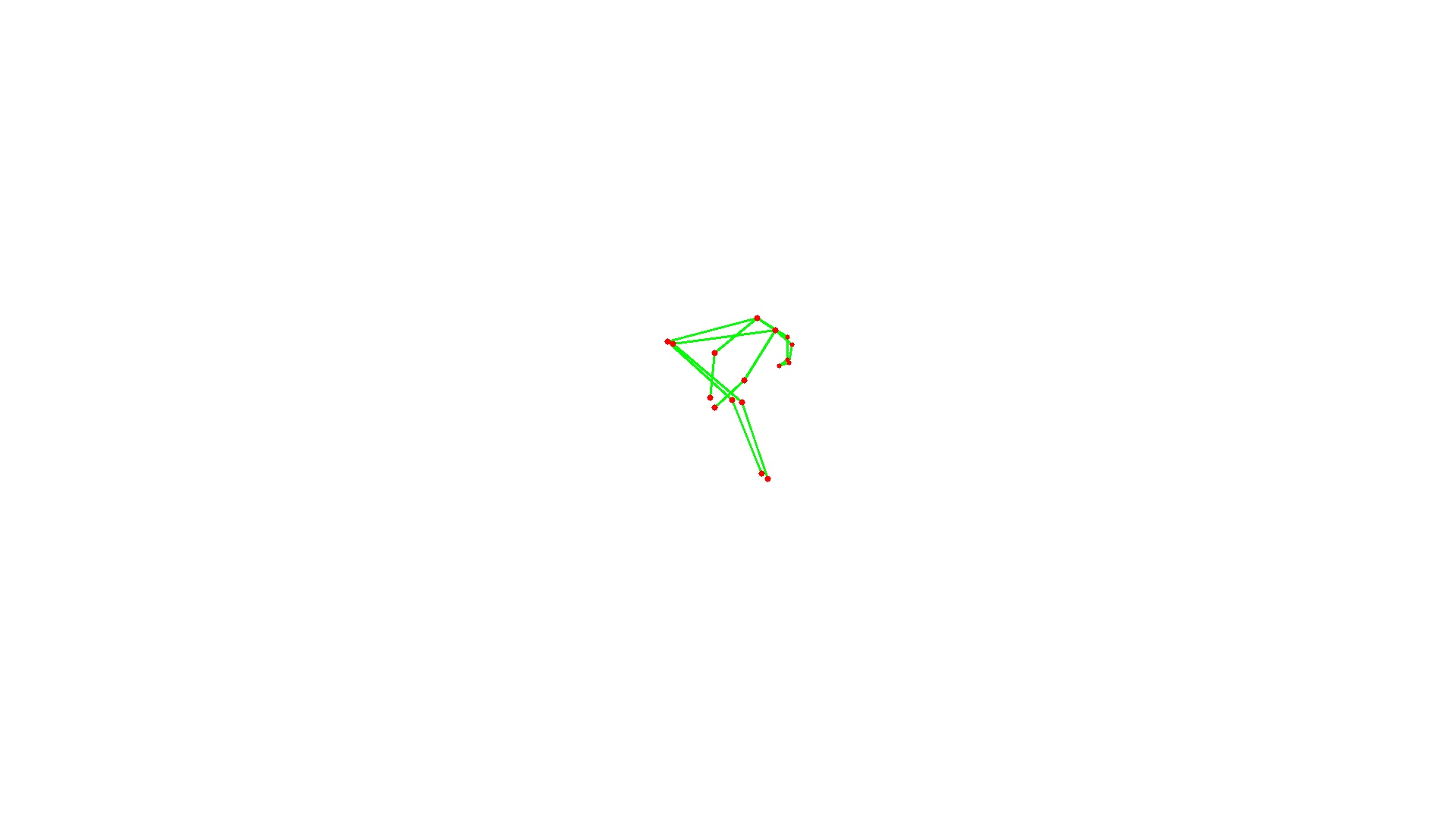}}\\[4pt]
  {\small Skeleton Render}

  \vspace{0.5em}
  {\small Figure \thefigure: MTL-AQA Preprocessing Examples.}
  \label{fig:mtlaqa_samples}
\end{center}


\clearpage
\section{Prompts}
\label{sec:appendix_prompts}

\tcbset{
  promptbox/.style={
    colback=gray!6,
    colframe=black!60,
    fonttitle=\bfseries\small,
    fontlower=\ttfamily\footnotesize,
    boxrule=0.4pt,
    arc=2pt,
    left=6pt, right=6pt, top=4pt, bottom=4pt,
    breakable
  },
  guidelinebox/.style={
    colback=gray!6,
    colframe=black!60,
    fonttitle=\bfseries\small,
    boxrule=0.4pt,
    arc=2pt,
    left=6pt, right=6pt, top=4pt, bottom=4pt
  }
}

We show all prompts used for each prompt engineering strategy and dataset.
For skeleton inputs, the phrase \textit{``a \{video~/~image\} of someone''} is replaced with
\textit{``a \{video~/~image\} showing skeleton joints (COCO17 format) of someone''}.
Placeholders in curly braces (e.g.\ \texttt{\{Action Name\}}) are filled at inference time with sample-specific values.

\subsection{Base Prompts}
\label{sec:prompts_base}

\begin{tcolorbox}[promptbox, title={Base Prompt — LLM-FMS}]
You are analyzing a \{Action Name\} image from a \{Camera View\} view. Answer the following questions about the person's form.

IMPORTANT: You must respond ONLY with a JSON object in the following format:
\{
  "\{Rule ID\}": "your\_answer",
\}

---

Question \{Rule ID\}: \{Rule Question\}
Options: \{Answer Options\}
Answer with one of the exact options listed.

...
\end{tcolorbox}

\medskip

\begin{tcolorbox}[promptbox, title={Base Prompt — EgoExo-Fitness}]
You are analyzing a video of someone performing the exercise: \{Action Name\}

Determine if the person is correctly following this instruction:
"\{Keypoint Statement\}"

IMPORTANT: Respond ONLY with "True" or "False" (nothing else).

Analyze the video carefully and provide your answer:
\end{tcolorbox}

\medskip

\begin{tcolorbox}[promptbox, title={Base Prompt — Fitness-AQA}]
You are analyzing a video of someone performing the exercise: \{Action Name\}

For each error type below, determine if the person is exhibiting that specific form error during the exercise.

IMPORTANT: Respond ONLY with a JSON object in the following format:
\{
  "\{Error Name\}": "True or False",
\}

Error types to check:

1. \{Error Name\}: \{Error Description\}

Analyze the video carefully and provide your JSON response:
\end{tcolorbox}

\medskip

\begin{tcolorbox}[promptbox, title={Base Prompt — FineFS}]
You are analyzing a video of someone executing a \{Action Name\} (\{Action Name\}).

For this execution, predict the Grade of Execution (GOE) score.

GOE Scale:

- Ranges from -5 (very poor) to 5 (exceptional)

- 0 indicates meeting basic requirements

- Positive GOE for good execution features (height, flow, control, technique)

- Negative GOE for errors, poor technique, or falls

IMPORTANT: Respond with ONLY a single numeric value from -5 to 5.

Analyze the video carefully and provide your response:
\end{tcolorbox}

\medskip

\begin{tcolorbox}[promptbox, title={Base Prompt — MTL-AQA}]
You are analyzing a video of someone executing a dive.

For this execution, predict the execution score.

Score Range:

- Ranges from 0 (very poor) to 10 (perfect)

- Higher scores indicate better execution quality

- Consider execution quality, body position, form, technique, and water entry

- Focus only on how well the dive is executed, not its difficulty

IMPORTANT: Respond with ONLY a single numeric value from 0 to 10.

Analyze the video carefully and provide your response:
\end{tcolorbox}

\subsection{Visual Grounding Prompts}
\label{sec:prompts_visual_grounding}

\begin{tcolorbox}[promptbox, title={Visual Grounding Prompt — LLM-FMS}]
You are analyzing a \{Action Name\} image from a \{Camera View\} view.

CRITICAL INSTRUCTIONS:

1. First, carefully examine the provided image in detail

2. Identify the person's body position, joint angles, and alignment

3. For each question below, look at the specific body parts mentioned in the image

4. Base your answer ONLY on what is actually visible in this specific image

5. Compare what you see against each available option to select the best match

RESPONSE FORMAT:
You must respond ONLY with a valid JSON object. No explanations, no additional text.
The JSON object must be in this exact format:
\{
  "\{Rule ID\}": "your\_answer",
\}

---

ANALYSIS QUESTIONS:
For each question, examine the relevant body parts in the image and select the option that matches your observation.

Question \{Rule ID\}: \{Rule Question\}
Look at the image and choose from: \{Answer Options\}
(Select the exact option that matches what you observe in the image)

...

---

Now examine the image and output your JSON response with answers based on what you see.
\end{tcolorbox}

\medskip

\begin{tcolorbox}[promptbox, title={Visual Grounding Prompt — EgoExo-Fitness}]
You are analyzing a video of someone performing the exercise: \{Action Name\}

CRITICAL INSTRUCTIONS:

1. First, carefully examine ALL frames in the provided video

2. Focus on the specific body parts and movements mentioned in the statement below

3. Observe how the person executes the movement throughout the video

4. Base your answer ONLY on what is actually visible in this specific video

---

STATEMENT TO VERIFY:
"\{Keypoint Statement\}"

Observe the video and determine: Is this instruction being correctly followed?
(Answer with the exact word that matches your observation)

---

RESPONSE FORMAT:
Respond with ONLY "True" or "False" (nothing else).
\end{tcolorbox}

\medskip

\begin{tcolorbox}[promptbox, title={Visual Grounding Prompt — Fitness-AQA}]
You are analyzing a video of someone performing the exercise: \{Action Name\}

CRITICAL INSTRUCTIONS:

1. First, carefully examine ALL frames in the provided video

2. Identify the person's body position, movement patterns, and form throughout

3. For each error type below, observe the specific body parts mentioned across frames

4. Base your answer ONLY on what is actually visible in this specific video

5. Determine if each specific form error is present at any point during the movement

RESPONSE FORMAT:
You must respond ONLY with a valid JSON object. No explanations, no additional text.
The JSON object must be in this exact format:
\{
  "\{Error Name\}": "True or False",
\}

---

ERROR DETECTION:
For each error type, examine the relevant body parts in the video and determine if the error is present.

Error 1 - \{Error Name\}: \{Error Description\}
Observe the video and determine: True or False

---

Now examine the video and output your JSON response with answers based on what you observe.
\end{tcolorbox}

\medskip

\begin{tcolorbox}[promptbox, title={Visual Grounding Prompt — FineFS}]
You are analyzing a video of someone executing a \{Action Name\} (\{Action Name\}).

CRITICAL INSTRUCTIONS:

1. First, carefully examine ALL frames in the provided video

2. Identify the execution quality, movement patterns, and form throughout

3. Observe technique, flow, control, and any errors or excellent features

4. Base your answer ONLY on what is actually visible in this specific video

5. Determine the Grade of Execution (GOE) score for this execution

RESPONSE FORMAT:
You must respond with ONLY a single numeric value. No explanations, no additional text.
Output only the numeric value from -5 to 5.

---

GOE SCALE:

- Ranges from -5 (very poor) to 5 (exceptional)

- 0 indicates meeting basic requirements

- Positive GOE for good execution features (height, flow, control, technique)

- Negative GOE for errors, poor technique, or falls

---

Now examine the video and output your numeric response based on what you observe.
\end{tcolorbox}

\medskip

\begin{tcolorbox}[promptbox, title={Visual Grounding Prompt — MTL-AQA}]
You are analyzing a video of someone executing a dive.

CRITICAL INSTRUCTIONS:

1. First, carefully examine ALL frames in the provided video

2. Identify the execution quality, body positions, and form throughout

3. Observe technique, water entry, rotation control, and any errors or excellent features

4. Base your answer ONLY on what is actually visible in this specific video

5. Determine the execution score for this dive

RESPONSE FORMAT:
You must respond with ONLY a single numeric value. No explanations, no additional text.
Output only the numeric value from 0 to 10.

---

SCORE RANGE:

- Ranges from 0 (very poor) to 10 (perfect)

- Higher scores indicate better execution quality

- Consider execution quality, body position, form, technique, and water entry

- Focus only on how well the dive is executed, not its difficulty

---

Now examine the video and output your numeric response based on what you observe.
\end{tcolorbox}

\subsection{Two-Step Prompts}
\label{sec:prompts_two_step}

\begin{tcolorbox}[promptbox, title={Two-Step Prompt — LLM-FMS}]
You are analyzing a \{Action Name\} image from a \{Camera View\} view.

TASK OVERVIEW:
You will analyze this exercise form image in two steps:

1. First, provide a detailed description of what you observe

2. Then, answer specific questions based on your observations

---

STEP 1 - DETAILED OBSERVATION:
Before answering any questions, carefully examine the image and describe:

- Overall body position and posture

- Head and neck alignment

- Shoulder position and alignment

- Spine curvature and torso position

- Hip position and alignment

- Knee position and alignment

- Foot placement and weight distribution

- Any notable angles or body segment relationships

Write your detailed observation here, then proceed to Step 2.

---

STEP 2 - ANSWER QUESTIONS:
Based on your detailed observation above, answer the following questions.
For each question, refer back to the specific body parts you described.

Question \{Rule ID\}: \{Rule Question\}
Choose from: \{Answer Options\}
(Select the exact option that matches what you observe in the image)

...

---

RESPONSE FORMAT:
Provide your response in this exact format:

OBSERVATION:
[Your detailed description of the image here]

ANSWERS:
\{
  "\{Rule ID\}": "your\_answer",
\}

Begin your analysis now.
\end{tcolorbox}

\medskip

\begin{tcolorbox}[promptbox, title={Two-Step Prompt — EgoExo-Fitness}]
You are analyzing a video of someone performing the exercise: \{Action Name\}

TASK OVERVIEW:
You will analyze this exercise video in two steps:

1. First, describe what you observe regarding the specific aspect mentioned

2. Then, determine if the statement is True or False

---

STATEMENT TO VERIFY:
"\{Keypoint Statement\}"

STEP 1 - OBSERVATION:
Describe what you observe in the video regarding this specific instruction.
Focus on the relevant body parts and movements mentioned.

STEP 2 - DETERMINATION:
Based on your observation, is the instruction being correctly followed?

---

RESPONSE FORMAT:
Provide your response in this exact format:

OBSERVATION:
[Your observation about the relevant body parts/movements]

ANSWER:
True or False

Begin your analysis now.
\end{tcolorbox}

\medskip

\begin{tcolorbox}[promptbox, title={Two-Step Prompt — Fitness-AQA}]
You are analyzing a video of someone performing the exercise: \{Action Name\}

TASK OVERVIEW:
You will analyze this exercise video in two steps:

1. First, provide a detailed description of what you observe

2. Then, determine if each form error is present based on your observations

---

STEP 1 - DETAILED OBSERVATION:
Before answering any questions, carefully examine the video and describe:

- Overall movement pattern and exercise technique

- Body position and posture throughout the movement

- Arm and hand positioning

- Leg and foot positioning

- Torso and spine alignment

- Joint angles and alignment (especially knees and elbows)

- Any notable form deviations or compensations

Write your detailed observation here, then proceed to Step 2.

---

STEP 2 - ERROR DETECTION:
Based on your detailed observation above, determine if each form error is present.
For each error, refer back to the specific aspects you described.

Error 1 - \{Error Name\}: \{Error Description\}
(Determine True or False based on your observation)

---

RESPONSE FORMAT:
Provide your response in this exact format:

OBSERVATION:
[Your detailed description of the video here]

ANSWERS:
\{
  "\{Error Name\}": "True or False",
\}

Begin your analysis now.
\end{tcolorbox}

\medskip

\begin{tcolorbox}[promptbox, title={Two-Step Prompt — FineFS}]
You are analyzing a video of someone executing a \{Action Name\} (\{Action Name\}).

TASK OVERVIEW:
You will analyze this execution in two steps:

1. First, provide a detailed description of what you observe

2. Then, determine the Grade of Execution (GOE) based on your observations

---

STEP 1 - DETAILED OBSERVATION:
Before assigning a score, carefully examine the video and describe:

- Overall movement pattern and execution quality

- Body position and posture throughout

- Technical aspects (form, positions, alignment)

- Speed, flow, and control

- Height/amplitude (for jumps and spins)

- Landing quality (for jumps)

- Any notable errors or excellent features

Write your detailed observation here, then proceed to Step 2.

---

STEP 2 - GOE PREDICTION:
Based on your detailed observation above, determine the GOE score.
Refer back to the specific aspects you described.

GOE Scale:

- Ranges from -5 (very poor) to 5 (exceptional)

- 0 indicates meeting basic requirements

- Positive GOE for good execution features (height, flow, control, technique)

- Negative GOE for errors, poor technique, or falls

---

RESPONSE FORMAT:
Provide your response in this exact format:

OBSERVATION:
[Your detailed description of the video here]

ANSWER:
[single numeric value from -5 to 5]

Begin your analysis now.
\end{tcolorbox}

\medskip

\begin{tcolorbox}[promptbox, title={Two-Step Prompt — MTL-AQA}]
You are analyzing a video of someone executing a dive.

TASK OVERVIEW:
You will analyze this execution in two steps:

1. First, provide a detailed description of what you observe

2. Then, determine the execution score based on your observations

---

STEP 1 - DETAILED OBSERVATION:
Before assigning a score, carefully examine the video and describe:

- Overall movement pattern and execution quality

- Body position and posture throughout the dive

- Technical aspects (form, positions, alignment)

- Rotation control and speed

- Water entry quality and splash

- Any notable errors or excellent features

Write your detailed observation here, then proceed to Step 2.

---

STEP 2 - SCORE PREDICTION:
Based on your detailed observation above, determine the execution score.
Refer back to the specific aspects you described.

Score Range:

- Ranges from 0 (very poor) to 10 (perfect)

- Higher scores indicate better execution quality

- Consider execution quality, body position, form, technique, and water entry

- Focus only on how well the dive is executed, not its difficulty

---

RESPONSE FORMAT:
Provide your response in this exact format:

OBSERVATION:
[Your detailed description of the video here]

ANSWER:
[single numeric value from 0 to 10]

Begin your analysis now.
\end{tcolorbox}

\subsection{Structured Reasoning Prompts}
\label{sec:prompts_structured}

\begin{tcolorbox}[promptbox, title={Structured Reasoning Prompt — LLM-FMS}]
You are analyzing an image of someone performing: \{Action Name\} (view: \{Camera View\})

SCORING QUESTIONS:

- \{Rule ID\}: \{Rule Question\} (Options: \{Answer Options\})

- ...

---

INSTRUCTIONS:
You must analyze this image using a structured reasoning process.
Follow the EXACT format below with the specified XML tags.

REQUIRED RESPONSE FORMAT:

\textless{}look\textgreater{}
Briefly describe what is happening in the image at a high level.
Focus on the overall posture and movement being performed.
\textless{}/look\textgreater{}

\textless{}decompose\textgreater{}
List the specific components you need to analyze to determine the correct scores.
These may include: specific body parts, angles between body parts, distances between body parts, alignment, or posture elements.
Be specific and relevant to the scoring criteria.
\textless{}/decompose\textgreater{}

\textless{}analyse\textgreater{}
For each component listed in \textless{}decompose\textgreater{}, provide a brief analysis:

- Component 1: [Your analysis of the first component]

- Component 2: [Your analysis of the second component]

- Continue for all listed components...
\textless{}/analyse\textgreater{}

\textless{}assess\textgreater{}
Based on your analysis, determine the correct answer for each question.
Explain your reasoning briefly for each.
\textless{}/assess\textgreater{}

<output>
\{
  "\{Rule ID\}": "\textless{}answer from options\textgreater{}",
\}
</output>

---

IMPORTANT RULES:

1. You MUST use all five tags in order: \textless{}look\textgreater{}, \textless{}decompose\textgreater{}, \textless{}analyse\textgreater{}, \textless{}assess\textgreater{}, <output>

2. The <output> tag must contain ONLY valid JSON with answers from the provided options

3. Do not skip any tags or change their order

4. Base your analysis ONLY on what is visible in the image

Begin your structured analysis now.
\end{tcolorbox}

\medskip

\begin{tcolorbox}[promptbox, title={Structured Reasoning Prompt — EgoExo-Fitness}]
You are analyzing a video of someone performing the exercise: \{Action Name\}

---

STATEMENT TO VERIFY:
"\{Keypoint Statement\}"

---

INSTRUCTIONS:
You must analyze this video using a structured reasoning process.
Follow the EXACT format below with the specified XML tags.

REQUIRED RESPONSE FORMAT:

\textless{}look\textgreater{}
Briefly describe what is happening in the video at a high level.
Focus on the overall movement and exercise being performed.
\textless{}/look\textgreater{}

\textless{}decompose\textgreater{}
List the specific components you need to analyze to verify the statement.
These may include: specific body parts, angles between body parts, distances between body parts, timing, or movement patterns.
Be specific and relevant to the statement being verified.
\textless{}/decompose\textgreater{}

\textless{}analyse\textgreater{}
For each component listed in \textless{}decompose\textgreater{}, provide a brief analysis:

- Component 1: [Your analysis of the first component]

- Component 2: [Your analysis of the second component]

- Continue for all listed components...
\textless{}/analyse\textgreater{}

\textless{}assess\textgreater{}
Based on your analysis, determine whether the statement is True or False.
Explain your reasoning briefly.
\textless{}/assess\textgreater{}

<output>
True or False
</output>

---

IMPORTANT RULES:

1. You MUST use all five tags in order: \textless{}look\textgreater{}, \textless{}decompose\textgreater{}, \textless{}analyse\textgreater{}, \textless{}assess\textgreater{}, <output>

2. The <output> tag must contain ONLY the word "True" or "False"

3. Do not skip any tags or change their order

4. Base your analysis ONLY on what is visible in the video

Begin your structured analysis now.
\end{tcolorbox}

\medskip

\begin{tcolorbox}[promptbox, title={Structured Reasoning Prompt — Fitness-AQA}]
You are analyzing a video of someone performing: \{Action Name\}

POSSIBLE ERRORS TO DETECT:

- \{Error Name\}: "\{Error Description\}"

---

INSTRUCTIONS:
You must analyze this video using a structured reasoning process.
Follow the EXACT format below with the specified XML tags.

REQUIRED RESPONSE FORMAT:

\textless{}look\textgreater{}
Briefly describe what is happening in the video at a high level.
Focus on the overall movement and exercise being performed.
\textless{}/look\textgreater{}

\textless{}decompose\textgreater{}
List the specific components you need to analyze to detect each error.
These may include: specific body parts, angles between body parts, distances between body parts, timing, or movement patterns.
Be specific and relevant to the errors being detected.
\textless{}/decompose\textgreater{}

\textless{}analyse\textgreater{}
For each component listed in \textless{}decompose\textgreater{}, provide a brief analysis:

- Component 1: [Your analysis of the first component]

- Component 2: [Your analysis of the second component]

- Continue for all listed components...
\textless{}/analyse\textgreater{}

\textless{}assess\textgreater{}
Based on your analysis, determine which errors are present (True) or absent (False).
Explain your reasoning briefly for each error.
\textless{}/assess\textgreater{}

<output>
\{
  "\{Error Name\}": "True or False",
\}
</output>

---

IMPORTANT RULES:

1. You MUST use all five tags in order: \textless{}look\textgreater{}, \textless{}decompose\textgreater{}, \textless{}analyse\textgreater{}, \textless{}assess\textgreater{}, <output>

2. The <output> tag must contain ONLY valid JSON with "True" or "False" for each error

3. Do not skip any tags or change their order

4. Base your analysis ONLY on what is visible in the video

Begin your structured analysis now.
\end{tcolorbox}

\medskip

\begin{tcolorbox}[promptbox, title={Structured Reasoning Prompt — FineFS}]
You are analyzing a video of someone executing a \{Action Name\} (\{Action Name\}).

GOE SCALE:

- Ranges from -5 (very poor) to 5 (exceptional)

- 0 indicates meeting basic requirements

- Positive GOE for good execution features (height, flow, control, technique)

- Negative GOE for errors, poor technique, or falls

---

INSTRUCTIONS:
You must analyze this video using a structured reasoning process.
Follow the EXACT format below with the specified XML tags.

REQUIRED RESPONSE FORMAT:

\textless{}look\textgreater{}
Briefly describe what is happening in the video at a high level.
Focus on the overall movement and execution being performed.
\textless{}/look\textgreater{}

\textless{}decompose\textgreater{}
List the specific components you need to analyze to assess the execution quality.
These may include: body positions, movement quality, technical elements, speed, flow, control, or any errors.
Be specific and relevant to evaluating this execution.
\textless{}/decompose\textgreater{}

\textless{}analyse\textgreater{}
For each component listed in \textless{}decompose\textgreater{}, provide a brief analysis:

- Component 1: [Your analysis of the first component]

- Component 2: [Your analysis of the second component]

- Continue for all listed components...
\textless{}/analyse\textgreater{}

\textless{}assess\textgreater{}
Based on your analysis, determine the appropriate GOE score.
Explain your reasoning briefly considering positive features and any errors.
\textless{}/assess\textgreater{}

<output>
[single numeric value from -5 to 5]
</output>

---

IMPORTANT RULES:

1. You MUST use all five tags in order: \textless{}look\textgreater{}, \textless{}decompose\textgreater{}, \textless{}analyse\textgreater{}, \textless{}assess\textgreater{}, <output>

2. The <output> tag must contain ONLY a single numeric value from -5 to 5

3. Do not skip any tags or change their order

4. Base your analysis ONLY on what is visible in the video

Begin your structured analysis now.
\end{tcolorbox}

\medskip

\begin{tcolorbox}[promptbox, title={Structured Reasoning Prompt — MTL-AQA}]
You are analyzing a video of someone executing a dive.

SCORE RANGE:

- Ranges from 0 (very poor) to 10 (perfect)

- Higher scores indicate better execution quality

- Consider execution quality, body position, form, technique, and water entry

- Focus only on how well the dive is executed, not its difficulty

---

INSTRUCTIONS:
You must analyze this video using a structured reasoning process.
Follow the EXACT format below with the specified XML tags.

REQUIRED RESPONSE FORMAT:

\textless{}look\textgreater{}
Briefly describe what is happening in the video at a high level.
Focus on the overall dive and execution being performed.
\textless{}/look\textgreater{}

\textless{}decompose\textgreater{}
List the specific components you need to analyze to assess the execution quality.
These may include: body positions, rotation control, technical elements, water entry, form, or any errors.
Be specific and relevant to evaluating this dive execution.
\textless{}/decompose\textgreater{}

\textless{}analyse\textgreater{}
For each component listed in \textless{}decompose\textgreater{}, provide a brief analysis:

- Component 1: [Your analysis of the first component]

- Component 2: [Your analysis of the second component]

- Continue for all listed components...
\textless{}/analyse\textgreater{}

\textless{}assess\textgreater{}
Based on your analysis, determine the appropriate execution score.
Explain your reasoning briefly considering positive features and any errors.
\textless{}/assess\textgreater{}

<output>
[single numeric value from 0 to 10]
</output>

---

IMPORTANT RULES:

1. You MUST use all five tags in order: \textless{}look\textgreater{}, \textless{}decompose\textgreater{}, \textless{}analyse\textgreater{}, \textless{}assess\textgreater{}, <output>

2. The <output> tag must contain ONLY a single numeric value from 0 to 10

3. Do not skip any tags or change their order

4. Base your analysis ONLY on what is visible in the video

Begin your structured analysis now.
\end{tcolorbox}

\subsection{Guideline Prompts}
\label{sec:prompts_guidelines}

The Positive and Negative Guidelines Prompts share the same structure, represented here with the placeholder \texttt{\{Guidelines\}}, which is substituted with best-form or worst-form guidelines depending on the variant.

\begin{tcolorbox}[promptbox, title={Guidelines Prompt — LLM-FMS}]
You are analyzing a \{Action Name\} image from a \{Camera View\} view. Answer the following questions about the person's form.

\{Guidelines\}

---

IMPORTANT: You must respond ONLY with a JSON object in the following format:
\{
  "\{Rule ID\}": "your\_answer",
\}

---

Question \{Rule ID\}: \{Rule Question\}
Options: \{Answer Options\}
Answer with one of the exact options listed.

...
\end{tcolorbox}

\medskip

\begin{tcolorbox}[promptbox, title={Guidelines Prompt — EgoExo-Fitness}]
You are analyzing a video of someone performing the exercise: \{Action Name\}

\{Guidelines\}

Determine if the person is correctly following this instruction:
"\{Keypoint Statement\}"

IMPORTANT: Respond ONLY with "True" or "False" (nothing else).

Analyze the video carefully and provide your answer:
\end{tcolorbox}

\medskip

\begin{tcolorbox}[promptbox, title={Guidelines Prompt — Fitness-AQA}]
You are analyzing a video of someone performing the exercise: \{Action Name\}

\{Guidelines\}

For each error type below, determine if the person is exhibiting that specific form error during the exercise.

IMPORTANT: Respond ONLY with a JSON object in the following format:
\{
  "\{Error Name\}": "True or False",
\}

Error types to check:

1. \{Error Name\}: \{Error Description\}

Analyze the video carefully and provide your JSON response:
\end{tcolorbox}

\medskip

\begin{tcolorbox}[promptbox, title={Guidelines Prompt — FineFS}]
You are analyzing a video of someone executing a \{Action Name\} (\{Action Name\}).

For this execution, predict the Grade of Execution (GOE) score.

GOE Scale:

- Ranges from -5 (very poor) to 5 (exceptional)

- 0 indicates meeting basic requirements

- Positive GOE for good execution features (height, flow, control, technique)

- Negative GOE for errors, poor technique, or falls

\{Guidelines\}

IMPORTANT: Respond with ONLY a single numeric value from -5 to 5.

Analyze the video carefully and provide your response:
\end{tcolorbox}

\medskip

\begin{tcolorbox}[promptbox, title={Guidelines Prompt — MTL-AQA}]
You are analyzing a video of someone executing a dive.

For this execution, predict the execution score.

Score Range:

- Ranges from 0 (very poor) to 10 (perfect)

- Higher scores indicate better execution quality

- Consider execution quality, body position, form, technique, and water entry

- Focus only on how well the dive is executed, not its difficulty

\{Guidelines\}

IMPORTANT: Respond with ONLY a single numeric value from 0 to 10.

Analyze the video carefully and provide your response:
\end{tcolorbox}

\subsection{Contrastive Prompts}
\label{sec:prompts_contrastive}

\begin{tcolorbox}[promptbox, title={Contrastive Prompt — LLM-FMS}]
You are comparing two images of someone performing \{Action Name\}.

Image 1 is shown first, Image 2 is shown second.

\{Contrastive Question\}

Answer with ONLY 1 or 2.
\end{tcolorbox}

\medskip

\begin{tcolorbox}[promptbox, title={Contrastive Prompt — EgoExo-Fitness}]
You are comparing two executions of \{Action Name\}.

Video 1 is shown first, Video 2 is shown second.

Which video better follows this instruction: "\{Keypoint Statement\}"

Answer with ONLY 1 or 2.
\end{tcolorbox}

\medskip

\begin{tcolorbox}[promptbox, title={Contrastive Prompt — Fitness-AQA}]
You are comparing two \{Action Name\} executions.

Video 1 is shown first, Video 2 is shown second.

Which execution has better form? Answer with ONLY 1 or 2.
\end{tcolorbox}

\medskip

\begin{tcolorbox}[promptbox, title={Contrastive Prompt — FineFS}]
You are comparing two \{Action Name\} executions.

Video 1 is shown first, Video 2 is shown second.

Which execution has better quality? Answer with ONLY 1 or 2.
\end{tcolorbox}

\medskip

\begin{tcolorbox}[promptbox, title={Contrastive Prompt — MTL-AQA}]
You are comparing two dive executions.

Video 1 is shown first, Video 2 is shown second.

Which execution has better quality? Answer with ONLY 1 or 2.
\end{tcolorbox}


\clearpage
\section{Exercise Guideline Examples}
\label{sec:appendix_guidelines}

We show one example of positive (best-form) and negative (worst-form) guidelines for each dataset, presented side by side.
These guidelines are injected into the \{Guidelines\} placeholder of the Guideline Prompts (Section~\ref{sec:appendix_prompts}).

\subsection{LLM-FMS (Deep Squat)}
\label{sec:guidelines_llmfms}

\begin{tcolorbox}[guidelinebox, width=\linewidth, title={Positive Guidelines — LLM-FMS}]
Key points for an ideal form in Deep Squat (Floor):

\textbullet{} The trunk should be parallel to the calf.

\textbullet{} The hip should be positioned lower than the knee on the vertical axis.

\textbullet{} The wrists should be positioned equal to the knees on the horizontal axis.
\end{tcolorbox}

\medskip

\begin{tcolorbox}[guidelinebox, width=\linewidth, title={Negative Guidelines — LLM-FMS}]
Common errors in Deep Squat (Floor):

\textbullet{} The trunk not being parallel to the calf.

\textbullet{} The hip being positioned higher than the knee on the vertical axis.

\textbullet{} The wrists being positioned to the left or right of the knees on the horizontal axis.
\end{tcolorbox}

\subsection{EgoExo-Fitness (Push-ups)}
\label{sec:guidelines_egoexo}

\begin{tcolorbox}[guidelinebox, width=\linewidth, title={Positive Guidelines — EgoExo-Fitness}]
Key points for an ideal form in Push-ups:

\textbullet{} The exercise should begin in a push-up position on the mat.

\textbullet{} The arms should be bent to lower the body towards the mat.

\textbullet{} The body should be descended until the elbows are slightly higher than the torso.

\textbullet{} The body should form a straight line when viewed from the side.

\textbullet{} The hands should be kept slightly wider than shoulder-width apart.

\textbullet{} The waist and back should be kept straight.

\textbullet{} The hands should be placed on the mat on both sides of the chest.

\textbullet{} The initial push-up position should be returned to.

\textbullet{} The arms should be stretched to push the body back up.
\end{tcolorbox}

\medskip

\begin{tcolorbox}[guidelinebox, width=\linewidth, title={Negative Guidelines — EgoExo-Fitness}]
Common errors in Push-ups:

\textbullet{} Not beginning in a push-up position on the mat.

\textbullet{} Not bending the arms to lower the body towards the mat.

\textbullet{} Not descending until the elbows are slightly higher than the torso.

\textbullet{} The body not forming a straight line when viewed from the side.

\textbullet{} Not keeping the hands slightly wider than shoulder-width apart.

\textbullet{} Not keeping the waist and back straight.

\textbullet{} Not placing the hands on the mat on both sides of the chest.

\textbullet{} Not returning to the initial push-up position.

\textbullet{} Not stretching the arms to push the body back up.
\end{tcolorbox}

\subsection{Fitness-AQA (Squat)}
\label{sec:guidelines_fitness}

\begin{tcolorbox}[guidelinebox, width=\linewidth, title={Positive Guidelines — Fitness-AQA}]
Key points for an ideal form in Squat:

\textbullet{} The knees should track over the toes without moving excessively forward beyond them during the descent.

\textbullet{} The knees should track over the toes during the descent and ascent phases without collapsing inward.
\end{tcolorbox}

\medskip

\begin{tcolorbox}[guidelinebox, width=\linewidth, title={Negative Guidelines — Fitness-AQA}]
Common errors in Squat:

\textbullet{} Knees tracking excessively forward beyond the toes during the descent, shifting weight to the front of the feet.

\textbullet{} Knees collapsing inward (valgus) during the descent or ascent phase, rather than tracking over the toes.
\end{tcolorbox}

\subsection{FineFS (Jump)}
\label{sec:guidelines_finefs}

\begin{tcolorbox}[guidelinebox, width=\linewidth, title={Positive Guidelines — FineFS}]
Key points for an ideal form in a jump:

\textbullet{} The take-off should be executed from the proper edge with controlled power.

\textbullet{} Sufficient height should be achieved to complete the required rotations.

\textbullet{} The body should be held in a tight rotational position in the air with arms drawn in close.

\textbullet{} The landing should be executed on a clean back outside edge.

\textbullet{} The landing knee should be bent adequately to absorb impact.

\textbullet{} The free leg should be extended behind on landing with proper turnout.

\textbullet{} Upper body posture should remain upright and controlled throughout the jump.

\textbullet{} Speed and flow should be maintained coming out of the landing.
\end{tcolorbox}

\medskip

\begin{tcolorbox}[guidelinebox, width=\linewidth, title={Negative Guidelines — FineFS}]
Common errors in a jump:

\textbullet{} Not executing the take-off from the proper edge with controlled power.

\textbullet{} Not achieving sufficient height to complete the required rotations.

\textbullet{} The body not being held in a tight rotational position in the air, with arms not drawn in close.

\textbullet{} Not landing on a clean back outside edge.

\textbullet{} Not bending the landing knee adequately to absorb impact.

\textbullet{} Not extending the free leg behind on landing with proper turnout.

\textbullet{} Upper body posture not remaining upright and controlled throughout the jump.

\textbullet{} Not maintaining speed and flow coming out of the landing.
\end{tcolorbox}

\subsection{MTL-AQA}
\label{sec:guidelines_mtlaqa}

\begin{tcolorbox}[guidelinebox, width=\linewidth, title={Positive Guidelines — MTL-AQA}]
Key points for an ideal form in a dive:

\textbullet{} The approach should be controlled with proper rhythm and consistent steps.

\textbullet{} The hurdle should provide adequate height and optimal distance from the board or platform.

\textbullet{} The take-off should be executed with proper timing and explosive power.

\textbullet{} The body should achieve the required position (pike, tuck, or straight) cleanly during flight.

\textbullet{} Rotations should be controlled and completed at the proper rate without under or over-rotating.

\textbullet{} The body should be fully straightened and aligned vertically before entry.

\textbullet{} The entry should be made at a vertical angle perpendicular to the water surface.

\textbullet{} The entry should create minimal splash with a clean ``rip'' technique.

\textbullet{} The toes should be pointed and legs together throughout the entire dive.

\textbullet{} The arms should be positioned properly, extended overhead and aligned with the body for entry.
\end{tcolorbox}

\medskip

\begin{tcolorbox}[guidelinebox, width=\linewidth, title={Negative Guidelines — MTL-AQA}]
Common errors in a dive:

\textbullet{} Not maintaining a controlled approach with proper rhythm and consistent steps.

\textbullet{} Not providing adequate height and optimal distance from the board or platform in the hurdle.

\textbullet{} Not executing the take-off with proper timing and explosive power.

\textbullet{} Not achieving the required position (pike, tuck, or straight) cleanly during flight.

\textbullet{} Rotations not being controlled or completed at the proper rate, resulting in under or over-rotation.

\textbullet{} The body not being fully straightened and aligned vertically before entry.

\textbullet{} Not making entry at a vertical angle perpendicular to the water surface.

\textbullet{} Creating excessive splash instead of a clean ``rip'' entry.

\textbullet{} The toes not being pointed and legs not together throughout the entire dive.

\textbullet{} The arms not being positioned properly, failing to extend overhead and align with the body for entry.
\end{tcolorbox}

\end{document}